%% file: main.tex
\documentclass[sigconf]{acmart}
\AtBeginDocument{%
  }

\usepackage{enumitem}
\usepackage{multirow}
\usepackage{soul}
\usepackage{balance}
\usepackage{xcolor}
\usepackage{subcaption}
\usepackage{algorithm}
\usepackage{algorithmic}
\usepackage{pifont}
\usepackage{makecell}
\renewcommand{\algorithmicrequire}{\textbf{Input:}}

\newenvironment{packed_lefty_item}{
\begin{itemize}[leftmargin=*]
\vspace{-2pt}
  \setlength{\itemsep}{0pt}
  \setlength{\parskip}{0pt}
  \setlength{\parsep}{0pt}
  \setlength{\topsep}{-10pt}
  \setlength{\partopsep}{0pt}
}{\end{itemize}\vspace{-2pt}}

\acmConference[]{Preprint}{Under review}{April 2025}

\begin{document}

\renewcommand\footnotetextcopyrightpermission[1]{}

\title{DiVE: Efficient Multi-View Driving Scenes Generation Based on Video Diffusion Transformer}

\author{Junpeng Jiang$^{1,2}$, Gangyi Hong$^{3,2}$, Miao Zhang$^{1*}$, Hengtong Hu$^{2}$, Kun Zhan$^{2}$, Rui Shao$^{1}$, Liqiang Nie$^{1}$}
\affiliation{%
  \institution{$^1$Harbin Institute of Technology, Shenzhen, $^2$Li Auto Inc., $^3$Tsinghua University}
  \city{Shenzhen}
  \country{China}
  \authornote{Corresponding author}
}
\email{jjunpeng1122@outlook.com, hgy23@mails.tsinghua.edu.cn, nieliqiang@gmail.com}
\email{{zhangmiao, shaorui}@hit.edu.cn, {zhankun, huhengtong}@lixiang.com}

\renewcommand{\shortauthors}{Junpeng Jiang et al.}

\begin{abstract}
Collecting multi-view driving scenario videos to enhance the performance of 3D visual perception tasks presents significant challenges and incurs substantial costs, making generative models for realistic data an appealing alternative. Yet, the videos generated by recent works suffer from poor quality and spatiotemporal consistency, undermining their utility in advancing perception tasks under driving scenarios. 
To address this gap, we propose DiVE, a diffusion transformer-based generative framework meticulously engineered to produce high-fidelity, temporally coherent, and cross-view consistent multi-view videos, aligning seamlessly with bird's-eye view layouts and textual descriptions.
Specifically, DiVE leverages a unified cross-attention and a SketchFormer to exert precise control over multimodal data, while incorporating a view-inflated attention mechanism that adds no extra parameters, thereby guaranteeing consistency across views.
Despite these advancements, synthesizing high-resolution videos under multimodal constraints introduces dual challenges: investigating the optimal classifier-free guidance (CFG) coniguration under intricate multi-condition inputs and mitigating excessive computational latency in high-resolution rendering---both of which remain underexplored in prior researches. To resolve these limitations, we introduce two technical innovations: (1) Multi-Control Auxiliary Branch Distillation, which streamlines multi-condition CFG selection while circumventing high computational overhead, and (2) Resolution Progressive Sampling, a training-free acceleration strategy that staggers resolution scaling to reduce high latency due to high resolution. These innovations collectively achieve a 2.62$\times$ speedup with minimal quality degradation. Evaluated on the nuScenes dataset, DiVE achieves state-of-the-art performance in multi-view video generation, yielding photorealistic outputs with exceptional temporal and cross-view coherence. By bridging the gap between synthetic data quality and real-world perceptual requirements, DiVE establishes a robust generative paradigm to catalyze significant advancements in 3D perception systems.

\end{abstract}

\keywords{Controllable Video Generation, Multi-View, Multi-Modal, Distillation}

\begin{teaserfigure}
    \vspace{-0.5cm}
    \begin{subfigure}{\textwidth}
        \includegraphics[width=\textwidth]{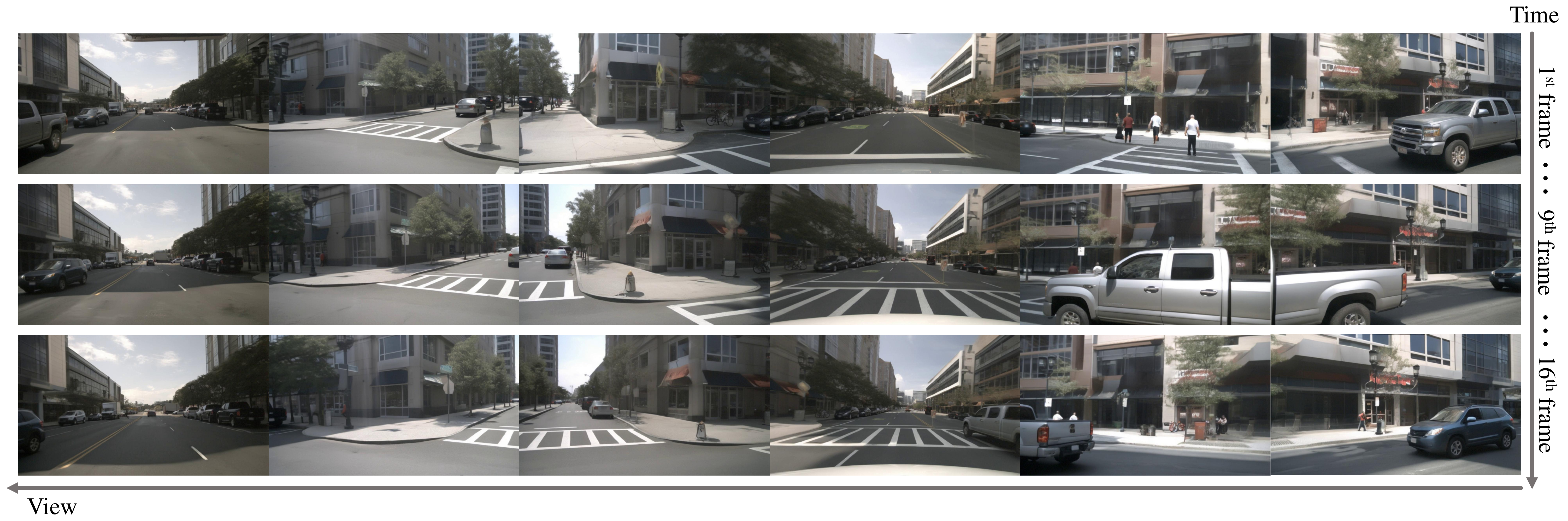}
        \vspace{-0.7cm}
        \caption{Multi-view video generation from DiVE.}
        \vspace{+0.3cm}
        \label{fig:main}
    \end{subfigure}
    \begin{subfigure}[b]{0.65\textwidth}
        \centering
        \includegraphics[width=\linewidth]{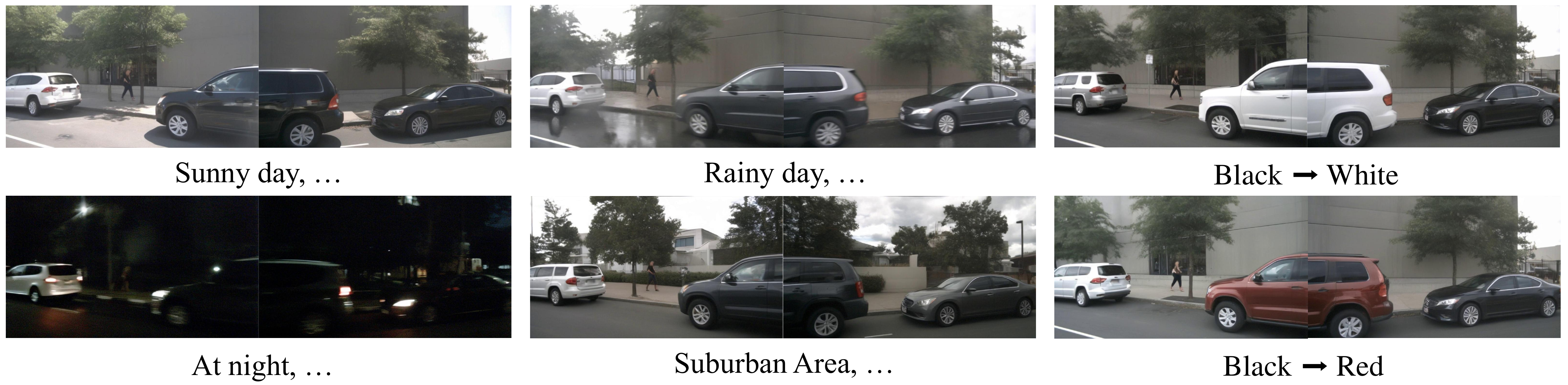}
        \vspace{-0.6cm}
        \caption{Scene and vehicle editing abilities of DiVE.}
        \label{fig:edit}
    \end{subfigure}   
    \hfill
    \begin{subfigure}[b]{0.34\textwidth}
        \centering
        \vspace{-0.5cm}
        \begin{tabular}{@{}l|c|c|c|c@{}}
            \toprule
            Method & CFG & mAP$\uparrow$ & NFE & SpeedUP$\uparrow$ \\ \midrule
            DiVE & w & 26.30 & 60 & 1$\times$  \\
            DiVE & w/o & 21.01 & 30 & 1.96$\times$    \\
            +MAD & w/o & 25.53 & 30 & 1.71$\times$   \\
            +RPS & w & 26.26 & 60 & 1.62$\times$   \\
            DiVE+ & w/o & 24.75 & 30 & \textcolor{red}{2.62$\times$}  \\ \bottomrule
        \end{tabular}
        \caption{DiVE with different acceleration strategies.}
        \label{tab:nfe}
    \end{subfigure}
    \vspace{-0.3cm}
    \caption{DiVE in qualitative visualizations and comparison of results under different acceleration techniques. (a) DiVE generates multi-view videos that exhibit strong realism, consistency, and controllability, making it an excellent simulator for controllably generating real-world driving scenarios under multi-modal conditions. (b) DiVE enables controllable video generation with varying attributes such as weather, time of day, location, and vehicle color. (c) CFG denots classifier-free guidance~\cite{ho2021classifier}. NFE denotes the number of function evaluations. We term DiVE incorporating the proposed Multi-Control Auxiliary Branch Distillation (MAD) and Resolution Progressively Sampling (RPS) as DiVE+, which significantly accelerate the inference without sacrificing performance.}
\end{teaserfigure}

\settopmatter{printacmref=false}

\maketitle

\input{section/1_intro}
\input{section/2_related}
\input{section/3_method}
\input{section/4_exper}
\input{section/5_conclusion}

\bibliographystyle{ACM-Reference-Format}
\bibliography{sample-base}

\appendix
\clearpage
{
\newpage
   \twocolumn[
    \centering
    \Large
    \textbf{DiVE: Efficient Multi-View Driving Scenes Generation Based on Video Diffusion Transformer}\\
    \vspace{0.5em}Supplementary Material \\
    \vspace{1.0em}
   ]
}
\input{section/appendix}
\end{document}

%% file: section/1_intro.tex
\section{Introduction}

3D visual perception tasks, such as 3D object detection and map segmentation, constitute foundational components of autonomous driving systems. 
Among existing approaches~\cite{huang2021bevdet,zhou2022cross,li2022bevformer,yang2023bevformer}, bird's-eye view (BEV) representations derived from multi-camera images have emerged as a dominant paradigm, owing to their ability to enable holistic environmental reasoning. Despite their effectiveness, these static-image-based frameworks inherently lack temporal context, which is critical for resolving time-sensitive attributes such as velocity~\cite{huang2022bevdet4d}. Multi-view video data, by contrast, offers rich spatiotemporal dynamics that can facilitate more precise perception in complex driving scenarios~\cite{wang2023exploring,liu2023sparsebev}. Nevertheless, the acquisition and annotation of diverse, real-world multi-view video datasets remain prohibitively expensive and labor-intensive. To address these limitations, synthetic data generation has emerged as a viable solution, with recent studies~\cite{swerdlow2024street,yang2023bevcontrol,gaomagicdrive,wen2024panacea,xie2025glad} demonstrating its efficacy in enhancing perception model performance. Motivated by these advancements, this paper focuses on the generation of controllable synthetic multi-view videos, designed to augment the training of 3D perception models in autonomous driving.

The fidelity of synthetic video data for autonomous driving perception hinges on three critical dimensions: annotation alignment, spatiotemporal consistency and spatial resolution. Among these, the first two factors are particularly vital for dynamic scene representation, as their degradation directly undermines the performance of downstream tasks such as object detection and tracking. Concurrently, insufficient spatial resolution obscures fine-grained scene elements, impairing precise environmental understanding~\cite{wang2023exploring,wen2024panacea+}. Despite these requirements, existing generation frameworks~\cite{gaomagicdrive,wen2024panacea} exhibit notable deficiencies---including temporal artifacts, viewpoint misalignment, and low resolution---that collectively impose fundamental limitations on the performance ceiling of perception models trained on such data. These issues, in part, from architectural constraints: while prior approaches~\cite{swerdlow2024street,yang2023bevcontrol,gaomagicdrive,wen2024panacea,wang2024driving,drivedreamer} rely on UNet-based architectures, state-of-the-art video generation models~\cite{wan2025,sora,yang2025cogvideox,kong2024hunyuanvideo} now predominantly adopt Diffusion Transformer (DiT)-based frameworks~\cite{peebles2023scalable}. However, existing DiT implementations, restricted by uni-modal text conditioning, remain incompatible with autonomous driving's multi-modal conditioning requirements, which necessitate simultaneous integration of four distinct representational modalities: linguistic descriptions (text), 2D spatial layouts (road sketches), 3D geometric structures (object instances), and temporal motion dynamics (camera parameters). The intrinsic heterogeneity of these modalities---each demanding specialized semantic parsing and spatiotemporal reasoning---poses significant challenges for conventional uni-modal architectures. To address this gap, we propose a novel framework that extends DiT’s capabilities to enable controllable multi-condition generation of multi-view driving scene videos.

In this paper, we propose \textbf{DiVE} (\textbf{Di}T-based \textbf{V}ideo Generation with \textbf{E}nhanced Multi-Modal Control), a novel framework that integrates DiT with multi-modal conditioning for controllable generation of multi-view driving scene videos. 
The architecture employs a unified cross-attention mechanism to jointly process textual scene descriptions, decomposed 3D objects, and camera parameters, achieving precise alignment of complex scenes and fine-grained control over foreground entities and motion trajectories. To enforce geometric consistency in BEV road layouts, inspired by~\cite{chen2024pixart,zhang2023adding}, we propose SketchFormer for sketch encoding. For multi-view coherence, we introduce view-inflated attention, an extra-parameter-free mechanism that propagates spatial dependencies across perspectives. Furthermore, a multi-scale training strategy is employed to enhance multi-scale feature learning and generation capabilities. 

Despite these advancements, high-resolution video generation under multi-modal conditioning presents substantial challenges. First, the combinatorial complexity of Classifier-free Guidance (CFG)~\cite{ho2021classifier} configurations across multiple conditions leads to intractable selection dilemmas. While omitting CFG could avoid this issue, it severely degrades output quality and controllability (Table~\ref{tab:nfe}). Moreover, theoretically optimal CFG configurations~\cite{brooks2023instructpix2pix} impose prohibitively high computational costs. Compounding these challenges, resolution requirements exponentially amplify inference-time computational demands for high-resolution outputs. To address these issues, we propose two complementary methodologies: \textbf{M}ulti-Control \textbf{A}uxiliary Branch \textbf{D}istillation (\textbf{MAD}) and \textbf{R}esolution \textbf{P}rogressively \textbf{S}ampling (\textbf{RPS}). MAD alleviates the configurational complexity and inefficiency of multi-condition CFG through efficient condition-specific auxiliary branches that distill guidance knowledge by fusing conditional inputs with the guidance scales. To amplify distillation efficacy, we introduce a novel mixed-control guidance training strategy, wherein auxiliary branches assimilate cross-condition guidance signals, thereby deepening their comprehension of own conditional guidance. RPS, a training-free inference strategy, tackles resolution-dependent inefficiency by leveraging multi-resolution generation capabilities. It performs initial and mid-stage inference at reduced resolutions, progressively transitioning to the target high resolution in later stages. This approach minimizes computational overhead while preserving output fidelity.

Our key contributions can be summarized as:
\begin{packed_lefty_item}
\item We propose DiVE, a novel DiT-based framework for multi-camera driving scene video generation, achieving state-of-the-art (SOTA) performance on the nuScenes~\cite{caesar2020nuscenes} dataset with a 36.7 FVD score reduction over prior SOTA methods. Furthermore, our synthetic data, characterized by high fidelity and multi-view consistency, significantly boosts downstream perception model performance.
\item We propose two synergistic methods: (1) Multi-Control Auxiliary Branch Distillation (MAD), which mitigates multi-condition CFG complexity via condition-specific auxiliary branches, enhanced by cross-condition knowledge distillation via mixed-control guidance training; and (2) Resolution Progressively Sampling (RPS), a training-free acceleration strategy, where the inference mode with progressively increasing resolution alleviates the computational burden required for the early-stage inference. Individually, MAD and RPS deliver 1.71$\times$ and 1.62$\times$ speedups, respectively. Their combination (DiVE+) achieves a 2.62$\times$ acceleration with minimal performance degradation.
\end{packed_lefty_item}

%% file: section/2_related.tex
\section{Related Work}
\noindent \textbf{Multi-View Generation for Driving Scenes.}
Multi-view driving scene generation relies on BEV layouts for synthesis. BEVGen~\citep{swerdlow2024street} generates street-view images autoregressively using spatial embeddings and camera bias. BEVControl~\citep{yang2023bevcontrol} employs attribute-specific controllers and cross-view-cross-element attention to ensure consistency. MagicDrive \citep{gaomagicdrive} leverages 3D scene encodings combined with cross-view attention for multi-perspective generation. Panacea~\citep{wen2024panacea} adopts a two-stage pipeline with decomposed 4D attention and a BEV-guided ControlNet for panoramic videos. Drive-WM~\citep{wang2024driving} integrates end-to-end planning with video generation. Despite progress, challenges in resolution, fidelity, and cross-view coherence remain, motivating our approach.

\noindent \textbf{Classifier-free Guidance Distillation.} Classifier-free guidance (CFG) has emerged as a powerful technique for improving sample quality across diverse generative paradigms, including class-conditional~\cite{peebles2023scalable,ma2024sit}, text-to-multimedia~\cite{podell2024sdxl,blattmann2023stable,gao2024catd}, and controllable diffusion models~\cite{li2023gligen,zheng2023layoutdiffusion}. However, its computational demands have spurred research into distilling guided sampling into efficient unguided surrogates. Existing methodologies, however, face critical limitations: Guided Distillation~\cite{meng2023distillation} irreversibly corrupts the original model’s score prediction capability via parameter overwriting and encodes knowledge exclusively at fixed guidance scales, precluding adaptive scaling during inference. Plug-and-Play~\cite{hsiao2024plug} enables scale conditioning via external modules but incurs latency penalties and fails to generalize to multi-modal conditions. DICE~\cite{zhou2025dice} proposes a lightweight enhancer combining null-text and text embeddings, offering a conceptual solution for multi-condition CFG distillation yet suffering significant performance degradation in multi-condition scenarios (Table ~\ref{tab:mad}). AGD~\cite{jensen2025efficient} employs adapter-based distillation but remains restricted to class-conditional generation. In constrast, MAD pioneers CFG distillation in complex multi-condition settings, achieving superior performance while maintaining computational efficiency through its innovative auxiliary branches and a mixed-control training paradigm.

%% file: section/3_method.tex
\section{Methodology}
\label{sec:method}

\begin{figure*}[htbp]
\centering
\vspace{-0.3cm}
\includegraphics[width=\linewidth]{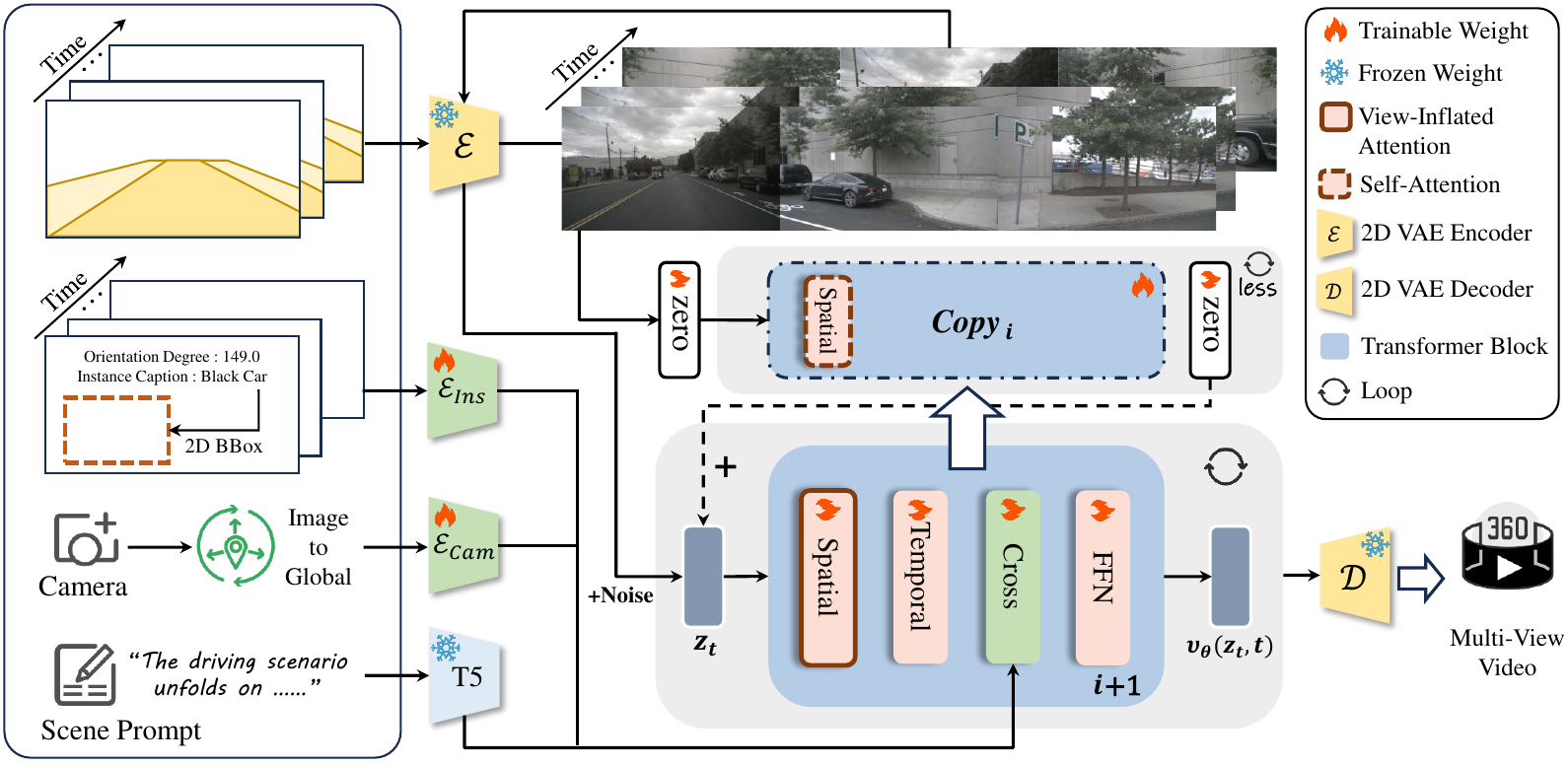}
\caption{Overview of DiVE for multi-view video generation. Our model encodes four inputs for controllable generation: scene description words for global context, camera information for motion control, bounding boxes that locate 3D objects placement, and road sketches for road conditions. Each block in DiVE consists of spatial attention, temporal attention, cross attention, and an MLP. Notably, the view-inflated attention, which enhances view consistency, is integrated into the spatial attention mechanism within the backbone network.}
\vspace{-0.3cm}
\label{fig:first}
\end{figure*}

In this section, we first delineate the mechanisms enabling DiVE's multi-modal controllability and cross-view consistency in Sec.~\ref{subsec:3.1}. Then Sec.~\ref{subsec:3.2} presents a distillation framework for classifier-free guidance under multi-condition settings. Finally, Sec.~\ref{subsec:3.3} introduces our training-free inference acceleration methodology.

\subsection{DiVE}
\label{subsec:3.1}
Figure~\ref{fig:first} outlines our model's architecture, adopting OpenSora 1.1~\cite{opensora} as the baseline. We extract latent features $z$ of multi-view video $x$ using a frozen LDM~\cite{rombach2022high} pre-trained VAE, then encode them with a 3D patch embedder to capture spatiotemporal dynamics.

\noindent \textbf{Unified Cross-Attention for Multi-Modal Conditioning.}
As shown in Figure~\ref{fig:first}, DiVE integrates multi-modal conditions via a unified cross-attention with three coordinated encoding pathways:
\begin{packed_lefty_item}
    \item \textbf{Linguistic Guidance.} We establish a dual-scale text conditioning system where scene-level context and instance-specific captions are encoded through T5~\cite{raffel2020exploring} and CLIP~\cite{radford2021learning} respectively. This yields semantic tokens $\mathcal{L}\in \mathbb{R}^{200\times d}$ and instance tokens $\mathcal{T}\in \mathbb{R}^{n_\text{ins}\times d}$, with $n_\text{ins}$ denoting the number of instances. 
    \item \textbf{Geometric Grounding.} Building on spatial disentanglement approach~\cite{yang2023bevcontrol}, 3D instances are projected into 2D space where bounding boxes $\mathcal{B}$ and orientation angles $\theta$ undergo Fourier encoding $\mathcal{F}$~\cite{mildenhall2021nerf}. These geometric features are fused with $\mathcal{T}$ through a parameterized blending MLP:
    \begin{equation}
        \mathcal{I}=\Phi\left (\mathcal{F}(\mathcal{B}),\mathcal{F}(\mathcal{\theta}),\mathcal{T}\right )~,
    \end{equation}
    where $\Phi$ denotes the fusion network.
    \item \textbf{Ego-motion Awareness.} To ensure cross-view consistency and motion coherence, we derive camera-aware embeddings $\mathcal{P}$ through coordinate transformation:
    \begin{equation}
        \mathcal{P} = \Phi\left ( \mathcal{F}\left (\begin{pmatrix}
              R&t \\
              0&1
            \end{pmatrix}\cdot \begin{pmatrix}
              K^{-1}&0 \\
              0&1
            \end{pmatrix} \right )\right )~,
    \end{equation}
    where rotation matrix $R\in \mathbb{R}^{3\times 3}$ and translation vector $t\in \mathbb{R}^{3\times 1}$ define the camera's pose in the global coordinate system. $K\in \mathbb{R}^{3\times 3}$ is the camera intrinsic matrix.
\end{packed_lefty_item}

The aggregated condition $\mathcal{C} = [\mathcal{L};\mathcal{I};\mathcal{P}]\in \mathbb{R}^{ \left ( 200 + n_\text{ins} + 1 \right ) \times d}$ establishes cross-modal correlations through cross-attention, harmonizing semantic, geometric, and dynamic constraints.

\noindent \textbf{SketchFormer for Road Guidance.}
Drawing from~\cite{chen2024pixart}, we introduce SketchFormer, a novel geometry-aware framework for road generation. Our method implements latent sketch guidance through a three-stage pipeline: (1) A pre-trained VAE compresses road sketches into disentangled embeddings; (2) A shared 3D patch embedder---identical to the primary network's---aligns these embeddings with the feature space; and (3) A cascade of 13 mirrored fusion cells, synchronized with the early layers of primary network, progressively blend sketch semantics via zero-initialized linear projections. This staged conditioning ensures spatial alignment while mitigating feature collision.

\noindent \textbf{View-Inflated Attention.}
In contrast to computationally heavy cross-view attention mechanisms~\cite{gaomagicdrive,wen2024panacea}, we propose view-inflated attention: a streamlined yet powerful approach that restructures input features from $V \times T \times H \times W \times C$ to $T \times (VHW) \times C$ before attention computation, where $V$, $T$, $H$, $W$, and $C$ denote views, frames, height, width and channel dimension, respectively. This parameter-free transformation implicitly enables cross-view interaction by treating $VHW$ as token length, achieving comparable multi-view consistency without introducing additional learnable weights. Notably, for SketchFormer, we intentionally bypass this reshaping step to preserve training stability and computational efficiency, as sketches provide inherent spatial constraints.

\noindent \textbf{Multi-Scale Training.} We propose a hierarchical training paradigm that progressively learns scale-aware representations through three distinct phases: (1) \textbf{Conditional Image Generation}, which establishes cross-modal grounding by mapping 3D constraints to multi-view images; (2) \textbf{Low-Resolution Video Training}, focusing on global spatiotemporal patterns and low-frequency features; and (3) \textbf{High-Resolution Video Refinement}, dedicated to capturing fine-grained visual details.
This multi-scale framework provides dual benefits: (1) implicit data augmentation through scale varia tion, and (2) native support for RPS (Sec.~\ref{subsec:3.3}).

\subsection{Multi-Control Auxiliary Branch Distillation}
\label{subsec:3.2}
In multi-condition scenarios, the combinatorial configurations of Classifier-free Guidance (CFG) exhibit remarkable diversity~\cite{liu2024stylecrafter,he2025cameractrl}.
Prior work~\cite{brooks2023instructpix2pix} theoretically offers an ideal solution but suffers from heavy computational overhead (\emph{e.g}., four function evaluations for three conditions) and suboptimal practical performance. To address these limitations, we introduce Multi-Control Auxiliary Branch Distillation (MAD), a novel framework that eliminates the need for CFG selection while enabling single-function-evaluation inference ( Figure~\ref{fig:mad}). MAD achieves this through two technical innovations:
\begin{figure}
    \centering
    \includegraphics[width=\linewidth]{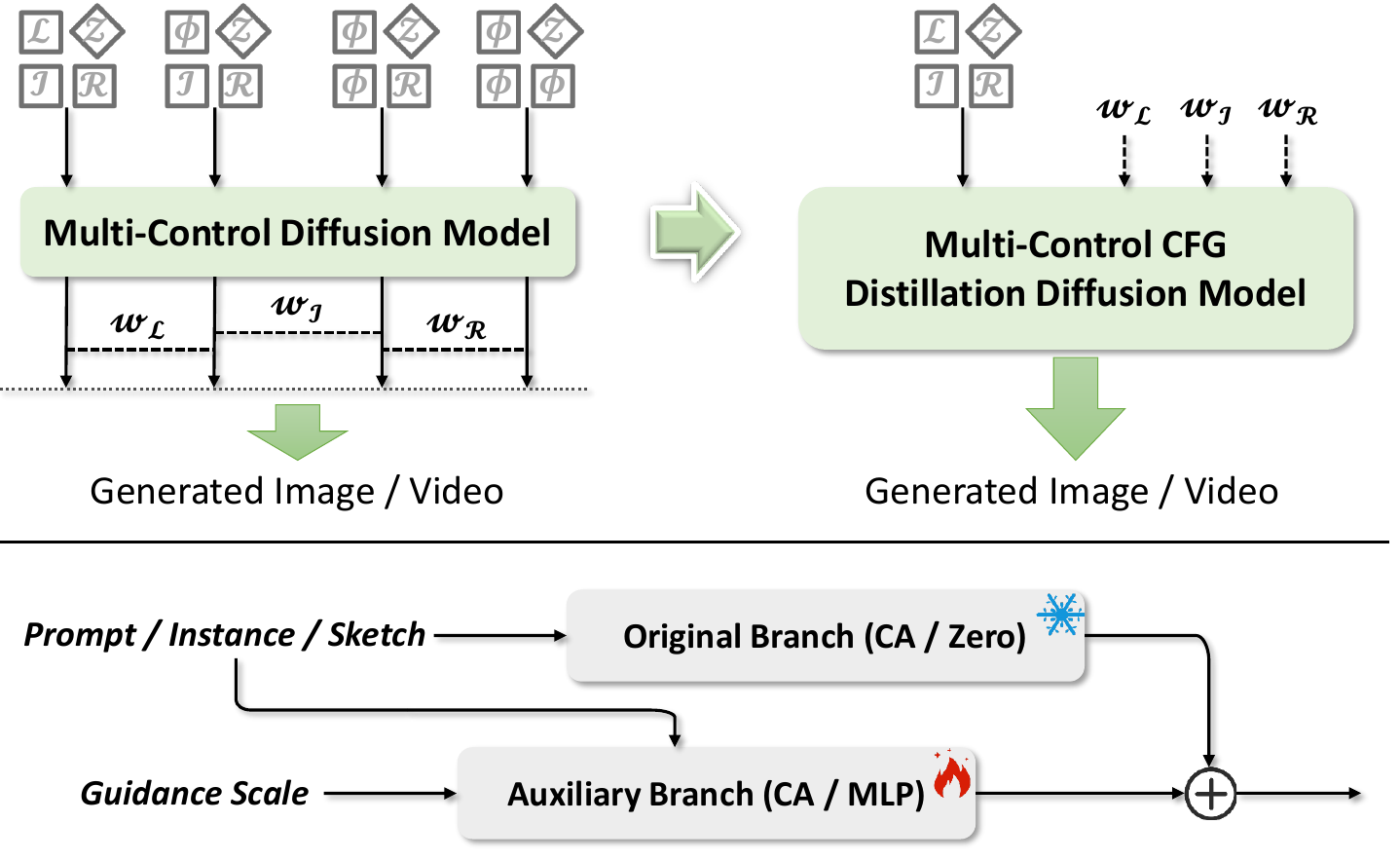}
    \caption{The overall process of Multi-Control Auxiliary Branch Distillation. CA denotes cross-attention.}
    \label{fig:mad}
    \vspace{-0.5cm}
\end{figure}
\begin{table*}[!t]
\centering
    \setlength{\tabcolsep}{4pt} % Adjust column spacing
    \renewcommand{\arraystretch}{1.2} % Improve row spacing
    \caption{Quantitative comparison of driving scenario generation methods evaluated on the nuScenes validation set. BEV segmentation and 3D object detection tasks use models pre-trained on nuScenes data. DiVE achieves top performance in all metrics. Even when integrated with MAD and RPS (DiVE+), superior performance is maintained. The best (\textbf{bold}) and second-best (\underline{underlined}) results are highlighted. Higher/lower metric values are preferred as indicated by $\uparrow/\downarrow$.}
    \vspace{-0.3cm}
    \label{tab:1}
    \begin{tabular}{@{}l|c|c|c|c|c|c|c|c|c@{}}
    \toprule
    \multirow{2}{*}{Method} & \multirow{2}{*}{Avenue} & \multirow{2}{*}{Resolution} & \multirow{2}{*}{FID$\downarrow$} & \multirow{2}{*}{FVD$\downarrow$} & \multirow{2}{*}{KPM(\%)$\uparrow$} & \multicolumn{2}{c|}{BEV segmentation} & \multicolumn{2}{c}{3D object detection} \\
    \cmidrule{7-8} \cmidrule{9-10}
     &  &  &  & &  & Road mIoU$\uparrow$ & Vehicle mIoU$\uparrow$ & mAP$\uparrow$ & NDS$\uparrow$ \\ 
    \midrule
    Real Data & - & - & - & - & - & 73.67 & 34.81 & 35.54 & 41.21 \\ \midrule
    BEVGen~\cite{swerdlow2024street} & RA-L'24 & 224$\times$400 & 25.54 & - & - & 50.20 & 5.89 & - & - \\
    BEVControl~\cite{yang2023bevcontrol} & arXiv'23 & - & 24.85 & - & - & 60.80 & 26.80 & 19.64 & - \\
    MagicDrive~\cite{gaomagicdrive} & ICLR'24 & 224$\times$400 & 16.20 & - & - & 61.05 & 27.01 & 12.30 & 23.32 \\ 
    MVPbev~\cite{liu2024mvpbev} & MM'24 & 256$\times$448 & 16.95 & - & - & 51.00 & - & - & - \\ \midrule
    DriveDreamer~\cite{drivedreamer} & ECCV'24 & 256$\times$448 & 26.80 & 353.2 & - & - & - & - & - \\
    DriveDreamer-2~\cite{zhao2024drivedreamer} & arXiv'24 & 256$\times$448 & 25.00 & 105.1 & - & - & - & - & - \\
    Panacea~\cite{wen2024panacea} & CVPR'24 & 256$\times$512 & 16.96 & 139.0 & 59.2 & 55.78 & 22.74 & 11.58 & 22.31 \\
    Drive-WM~\cite{wang2024driving} & CVPR'24 & 192$\times$384 & 15.80 & 122.7 & 45.8 & 65.07 & 27.19 & 20.66 & - \\
    Glad~\cite{xie2025glad} & ICLR'25 & 256$\times$512 & 12.57 & 207.0 & - & - & - & - & - \\ \midrule
    DiVE & - & 480$\times$854 & \textbf{7.14} & \textbf{68.4} & \textbf{73.2} & \textbf{68.16} & \textbf{30.50} & \textbf{25.75} & \textbf{33.61} \\
    DiVE+ & - & 480$\times$854 & \underline{10.68} & \underline{93.2} & \underline{63.0} & \underline{66.95} & \underline{30.42} & \underline{23.68} & \underline{31.97} \\ \bottomrule
    \end{tabular}
\end{table*}
\begin{figure}
    \centering
    \includegraphics[width=\linewidth]{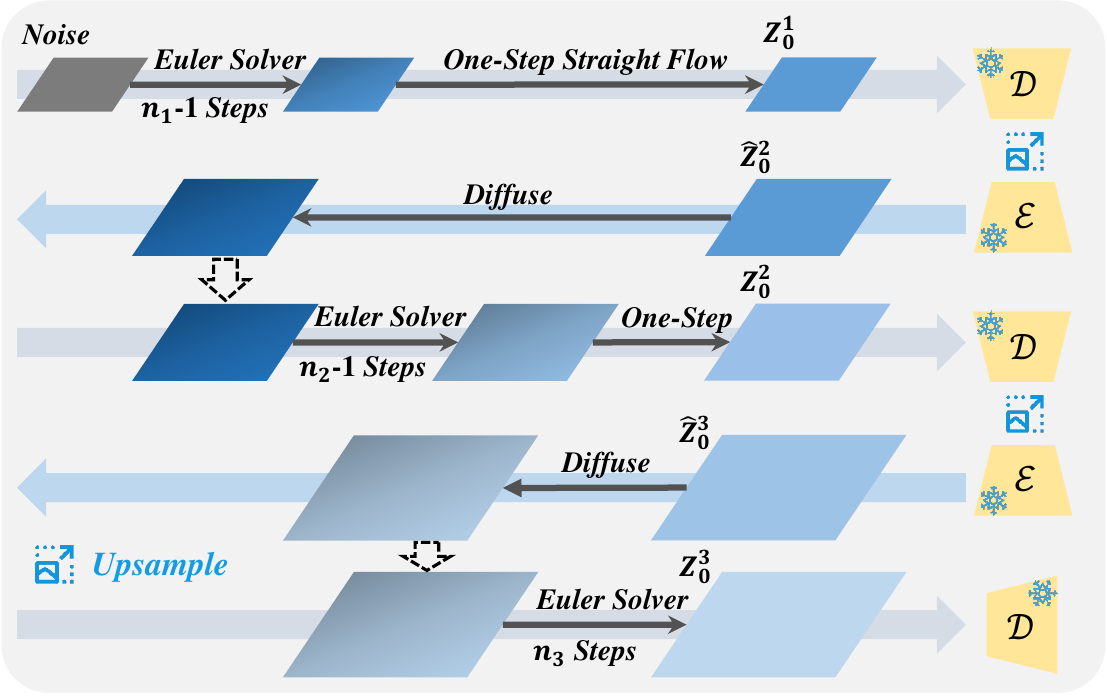}
    \caption{The overall process of Resolution Progressively Sampling. Larger quadrilaterals represent higher resolutions, and deeper colors indicate more noisy regions.}
    \label{fig:second}
    \vspace{-0.5cm}
\end{figure}
\begin{packed_lefty_item}
    \item \textbf{Auxiliary Branch.} MAD's auxiliary branch (Figure~\ref{fig:mad}) adopts a hybrid design: cross-attention modules for text/object guidance and MLPs for road sketch conditioning, which balances representational capacity and computational efficiency. Cross-attention---serving as DiVE's principal pathway for conditional information injection---provides a low-overhead solution (4.24\% of total runtime) for auxiliary knowledge distillation. For road sketches, we replace cross-attention with an efficient MLP as the auxiliary branch of zero-linear which bridges the connection between SketchFormer and the main network. This design stems from our observation that spatial attention mechanisms in SketchFormer outweigh cross-attention costs. Retaining cross-attention for road sketches would introduce redundant computations, whereas the proposed MLP preserves critical knowledge distillation with minimal overhead.
    \item \textbf{Mixed-Control Guidance Training.} Table~\ref{tab:mad} reveals that inter-condition non-interaction during auxiliary branch training impairs the model's ability to discern CFG information disparities across conditional-unconditional combinations. To resolve this, MAD's training strategy isolates condition-specific guidance signals via control conditions mixed distillation. Each training iteration designates at least one condition for knowledge transfer, requiring only two forward passes per sampling step---agnostic to the number of conditions distilled. The conditional CFG branch processes all conditions simultaneously, while the unconditional branch strategically nullifies target conditions during distillation. Algorithm~\ref{alg:alg} details this procedure.
\end{packed_lefty_item} 

These dual innovations form a synergistic relationship: the former establishes a robust guidance framework, while the latter refines the former’s understanding of conditional guidance disparities. This symbiotic relationship culminates in the MAD architecture, which harmonizes efficiency and performance. As demonstrated in Figure~\ref{fig:accab}, integrating MAD into DiVE preserves computational efficiency comparable to its CFG-disabled variant while achieving significant improvements in output quality.

\begin{figure}
    \centering
    \vspace{-0.5cm}
    \includegraphics[width=\linewidth]{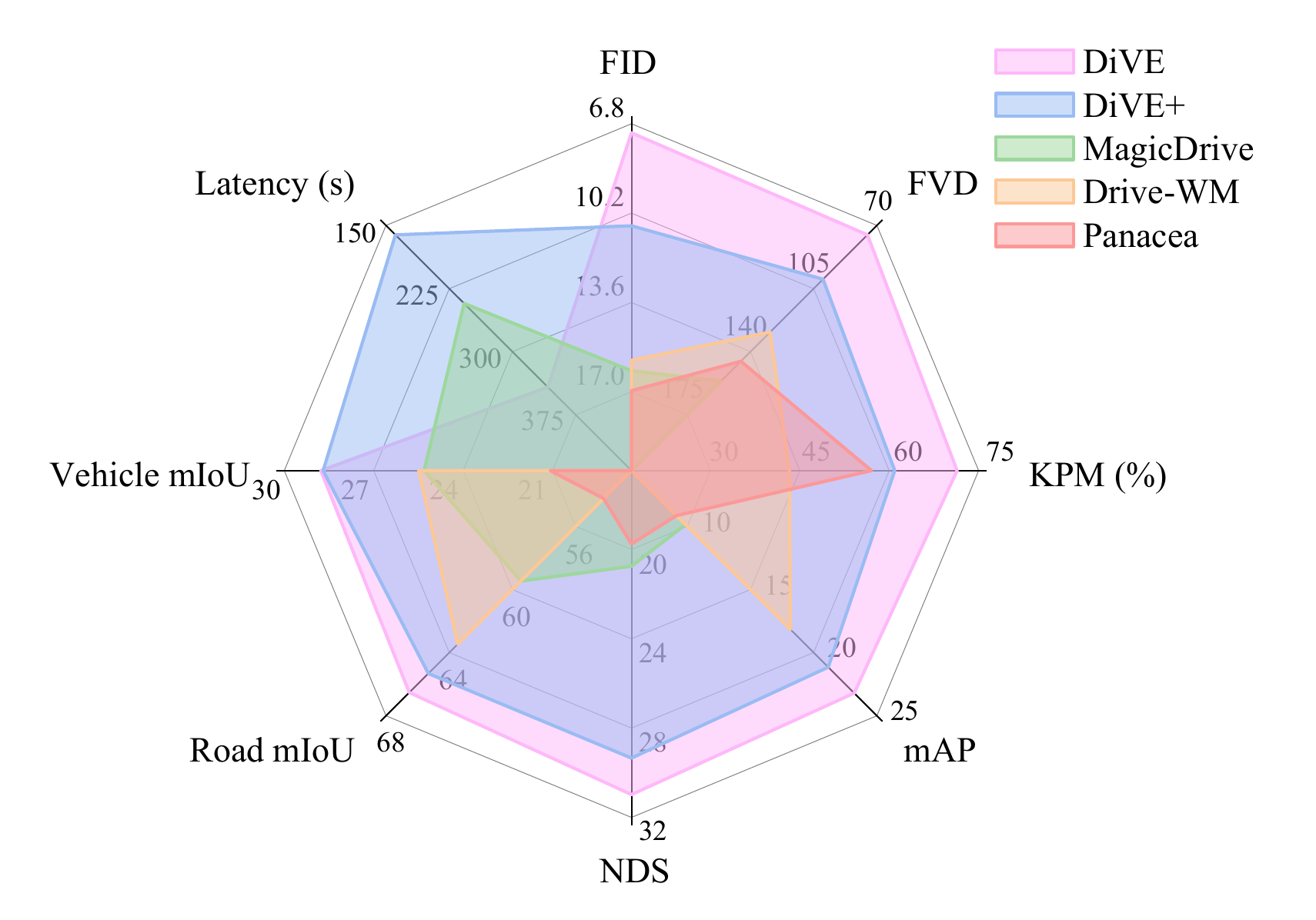}
    \vspace{-0.8cm}
    \caption{Visualization of quantitative comparison.}
    \vspace{-0.5cm}
    \label{fig:leida}
\end{figure}

\begin{figure*}[t]
    \centering
    \vspace{-0.3cm}
    \includegraphics[width=0.99\linewidth]{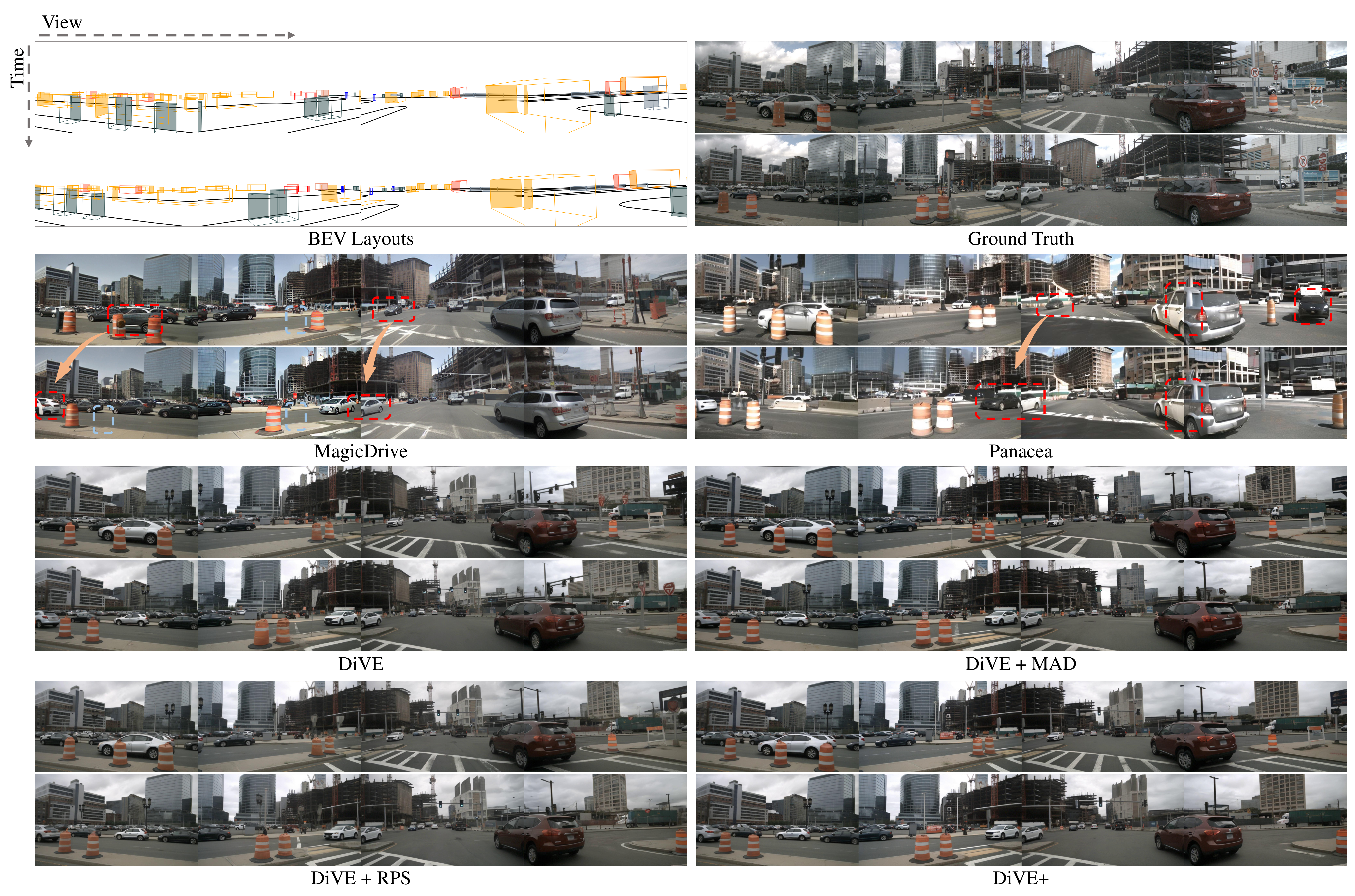}
    \vspace{-0.3cm}
    \caption{Qualitative comparison of DiVE with MagicDrive and Panacea. We use dashed boxes to highlight some of the noticeable issues in MagicDrive and Panacea, and arrows to indicate the changes in vehicle positions over time. In contrast, DiVE demonstrates superior realism, temporal and cross-view consistency, and controllability, both before and after applying RPS or MAD.}
    \label{fig:qualitative_cmp}
    \vspace{-0.5cm}
\end{figure*}

\subsection{Resolution Progressively Sampling}
\label{subsec:3.3}
To mitigate computational pressure in high-resolution video generation, we propose Resolution Progressively Sampling (RPS), a training-free inference acceleration strategy (Figure~\ref{fig:second}). It leverages the multi-resolution capability of DiVE via two key innovations: 
\begin{packed_lefty_item}
    \item \textbf{Progressive Resolution Scaling.} Inspired by~\cite{du2024demofusion, wu2024megafusion}, we employ a multi-stage sampling framework that iteratively refines the video from low-resolution (\emph{e.g}., 240p) latent $z_1$ to high-resolution (\emph{e.g}., 480p) output $x$, with $s$ stages and $n_k$ steps per stage $(N=\sum_{k=1}^{s}n_k)$. Within each stage $S_k$, sampling timesteps decrease from $t_{n_{k}}^{k}$ to $t_{1}^{k}$. 
    Between stages, after completing the first $n_{k}-1$ steps, a one-step straight flow at $t_{1}^{k}$ estimates $z_{0}^{k}$, which is decoded by the VAE into a clear video at the current resolution. This video is upsampled and encoded into the initial state of the next stage via diffusion, creating a cyclic progression from low to high resolution until the target resolution $h_s \times w_s$ is achieved.
    \item \textbf{Resolution-Aware Timestep Shift.} Motivated by~\cite{esser2024scaling,hoogeboom2023simple}, higher resolutions require greater noise to effectively disrupt the signal. The relationship between $t_{n_{k+1}}^{k+1}$ and $t_{1}^k$ follows a shift function:
        \begin{equation}
        t_{n_{k+1}}^{k+1}=\frac{t_{1}^k\sqrt{(h_{k+1} w_{k+1})/(h_{k} w_{k}) } }{1+t_1^k\left(\sqrt{(h_{k+1} w_{k+1})/(h_{k} w_{k})} -1\right)}  ~,
        \end{equation}
    ensuring increased noise at the $k+1$ stage.
\end{packed_lefty_item}

As shown in Figure~\ref{fig:accab} and Table~\ref{tab:4}, RPS achieves 1.62$\times$ faster inference versus baseline, while enhancing visual fidelity through log-SNR consistency $\text{log}\frac{h_{k} w_{k}}{h_{k+1} w_{k+1}} $~\cite{hoogeboom2023simple}.

\begin{table*}
\vspace{-0.1cm}
    \centering
    \setlength{\tabcolsep}{4pt}
    \renewcommand{\arraystretch}{1.2}
    \captionsetup{width=.95\linewidth}
    \caption{
    Comparison of support for StreamPETR. The results are reported on the nuScenes validation set. The experimental results on Panacea is re-implemented.
    }
    \vspace{-0.2cm}
    \label{tab:2}
    \begin{tabular}{@{}l|c|c|c|c|c|c|c|c|c|c@{}}
    \toprule
    Method & Resolution & Real & Generated & mAP$\uparrow$ & mATE$\downarrow$ & mASE$\downarrow$ & mAOE$\downarrow$ & mAVE$\downarrow$ & mAAE$\downarrow$ & NDS$\uparrow$ \\ \midrule
     & & $\checkmark$ & - & 34.6 & 66.7 & 27.4 & 59.2 & 29.0 & 20.4 & 47.0 \\
    Panacea~\cite{wen2024panacea} & 256$\times$512 & - & $\checkmark$ & 22.8 & 79.2 & 28.6 & 70.3 & 47.1 & 22.9 & 36.6 \\
     & & $\checkmark$ & $\checkmark$ & 36.3 \textcolor{blue}{(+1.7\%)} & 65.6 & 27.1 & 51.9 & 29.0 & 18.6 & 48.9 \textcolor{blue}{(+1.9\%)} \\ \midrule
     Glad~\cite{xie2025glad} & 256$\times$512 & $\checkmark$ & $\checkmark$ & 37.1 \textcolor{blue}{(+2.5\%)} & - & - & - & - & - & 49.2 \textcolor{blue}{(+2.2\%)} \\ \midrule
     & & $\checkmark$ & - & 38.0 & 66.7 & 26.8 & 52.7 & 32.1 & 20.9 & 49.0 \\
    DiVE & 480$\times$854 & - & $\checkmark$ & 27.0 & 79.1 & 27.6 & 55.5 & 48.3 & 20.9 & 40.4 \\
     & & $\checkmark$ & $\checkmark$ & \textbf{40.9} \textcolor{blue}{(+2.9\%)} & \textbf{62.7} & \textbf{26.4} & \textbf{48.6} & \textbf{28.5} & \textbf{17.9} & \textbf{52.0} \textcolor{blue}{(+3.0\%)} \\ \bottomrule
    \end{tabular}
    \vspace{-0.3cm}
\end{table*}
\begin{table}
    \centering
    \setlength{\tabcolsep}{4pt}
    \renewcommand{\arraystretch}{1.2}
    \caption{
    Ablation on the different spatial attention types.
    }
    \vspace{-0.2cm}
    \label{tab:3}
    \begin{tabular}{@{}c|c|c|c|c@{}}
    \toprule
    Attention Type & FVD$\downarrow$ & KPM(\%)$\uparrow$ & Object mAP$\uparrow$ & Map mIoU$\uparrow$ \\ \midrule
    Self & 93.14 & 58.4 & 24.57 & 35.87 \\
    Left-Self-Right & 86.15 & \textbf{76.6} & 25.88 & 36.97 \\
    \textbf{View-Inflated} & \textbf{86.13} & 73.2 & \textbf{26.30} & \textbf{37.41} \\ \bottomrule
    \end{tabular}
    \vspace{-0.3cm}
\end{table}

%% file: section/4_exper.tex
\section{Experiments}

\subsection{Experimental Setups}

\noindent \textbf{Dataset.}
We conduct experiments on the nuScenes~\cite{caesar2020nuscenes} dataset, a publicly available 3D perception dataset for driving scenarios. The nuScenes dataset comprises 700 video sequences for training and 150 for validation. Each sequence is recorded at 12 Hz and lasts approximately 20 seconds, with annotations provided at 2 Hz. For high-frame-rate generation, we utilize the 12 Hz interpolated annotations provided by W-CODA\footnote{W-CODA's homepage:~\href{https://coda-dataset.github.io/w-coda2024/}{https://coda-dataset.github.io/w-coda2024/}} for both training and evaluation.

\noindent \textbf{Quality Metrics.} To evaluate the quality of the generated video, we utilize two primary metrics: the frame-wise Fréchet Inception Distance (FID)~\cite{heusel2017gans} and the Fréchet Video Distance (FVD)~\cite{unterthiner2018towards}. FID assesses the quality of individual frames, whereas FVD evaluates both the quality and temporal consistency of the video. Complementing these assessments, we further use a pre-trained matching model~\cite{sun2021loftr} to compute the Key Points Matching (KPM) score~\cite{wang2024driving} to analyze multi-view consistency.

\noindent \textbf{Controllability Metrics.} To assess controllability, we employ CVT~\cite{zhou2022cross} and BEVFusion~\cite{liu2023bevfusion} to conduct a quantitative analysis of two perception tasks—BEV segmentation and 3D object detection. We generate the corresponding data based on the annotations from the validation set and evaluate performance using a model pretrained on real data. Additionally, we generate data on the training set and utilize the video-based perception method StreamPETR~\cite{wang2023exploring} to evaluate the effectiveness of DiVE in augmenting data.

\subsection{Implementation Details.}
Our implementation is based on the OpenSora 1.1~\cite{opensora} codebase, initialized with pretrained weights. The multi-scale training process is carried out on 8 NVIDIA A800 GPUs, with each of the three phases comprising 20$k$, 30$k$ and 80$k$ iterations, respectively. For inference, we utilize rectified flow~\cite{liuflow} with a classifier-free guidance~\cite{ho2021classifier} scale of 2.0, performing 30 sampling steps to generate videos at a resolution of 480p (480$\times$854) and 16 frames. For MAD, during each training iteration, a guidance scale value is uniformly sampled fron the continous interval $[1,8]$, and the distilled model is trained for a total of 8$k$ iterations. Our RPS acceleration strategy conducts 10 sampling steps at resolutions of 240p (240$\times$426), 360p (360$\times$640), and 480p (480$\times$854) sequentially. 

\subsection{Main Results}

\noindent \textbf{Quantitative Results.} 
Table~\ref{tab:1} compares DiVE’s performance against prior methods on the nuScenes validation. Our model achieves state-of-the-art FID (7.14), FVD (68.4) and KPM (73.2) scores, surpassing existing multi-view image/video generation models. This highlights DiVE's unique advantage in maintaining cross-view and temporal consistency while preserving frame-level quality. Control precision analysis further demonstrates its strengths: BEV segmentation performance (Road and Vehicle mIoU) exhibits strong alignment with real data distributions, while 3D detection metrics (mAP and NDS) validate precise geometric correspondence. Despite minor performance reductions when applying MAD and RPS, DiVE maintains superior performance across all metrics. As illustrated in Figure \ref{fig:leida}, DiVE achieves significant improvements in nearly all metrics (except latency), while its accelerated variant DiVE+ comprehensively surpasses previous approaches. These quantitative results confirm DiVE's dual capability to simultaneously optimize photorealism and geometric fidelity.

\begin{table}
    \caption{Ablation of Multi-Control Auxiliary Branch Distillation. PnP denotes Plug-and-Play Diffusion Distillation~\cite{hsiao2024plug}. PnP and DICE~\cite{zhou2025dice} are re-implemented with the same training iterations as our method.}
    \vspace{-0.2cm}
    \label{tab:mad}
    \centering
    \setlength{\tabcolsep}{4pt}
    \renewcommand{\arraystretch}{1.2}
    \begin{tabular}{@{}c|c|c|c|c@{}}
    \toprule
    Method & FVD$\downarrow$ & Object mAP$\uparrow$ & Map mIoU$\uparrow$ & Latency(s)$\downarrow$ \\ \midrule
    PnP & 169.20 & 24.72 & 36.85 & 194.7 \\
    DICE & 174.63 & 18.97 & 32.59 & \textbf{157.3} \\ \midrule
    Strategy 1 & 134.60 & 24.68 & 36.89 & 179.5 \\
    Strategy 2 & 141.41 & 24.01 & 36.13 & 179.5 \\ \midrule
    \textbf{MAD} & \textbf{100.18} & \textbf{25.53} & \textbf{37.47} & 179.5 \\ \bottomrule
    \end{tabular}
    \vspace{-0.3cm}
\end{table}

\begin{table}
    \caption{Ablation of Resolution Progressively Sampling. TS represents resolution-aware timestep shift.}
    \vspace{-0.1cm}
    \label{tab:4}
    \centering
    \setlength{\tabcolsep}{4pt}
    \renewcommand{\arraystretch}{1.2}
    \begin{tabular}{@{}c|c|c|c|c|c@{}}
    \toprule
    Choices & TS & FVD$\downarrow$ & Object mAP$\uparrow$ & Map mIoU$\uparrow$ & Latency(s)$\downarrow$ \\ \midrule
    10-10-10 & - & 103.92 & 25.29 & 37.60 & 189.7 \\
    0-0-30 & - & 86.13 & 26.30 & 37.41 & 307.1 \\ \midrule
    20-5-5 & $\checkmark$ & 119.62 & 24.54 & 36.72 & 134.4 \\
    5-5-20 & $\checkmark$ & 89.64 & 26.26 & 37.92 & 252.7 \\
    10-10-10 & $\checkmark$ & 97.74 & 26.26 & 37.68 & 189.7 \\ \bottomrule
    \end{tabular}
    \vspace{-0.4cm}
\end{table}

\begin{figure*}[t]
\vspace{-0.8cm}
    \centering
    \begin{minipage}[t]{0.34\textwidth}
        \centering
        \includegraphics[width=\textwidth]{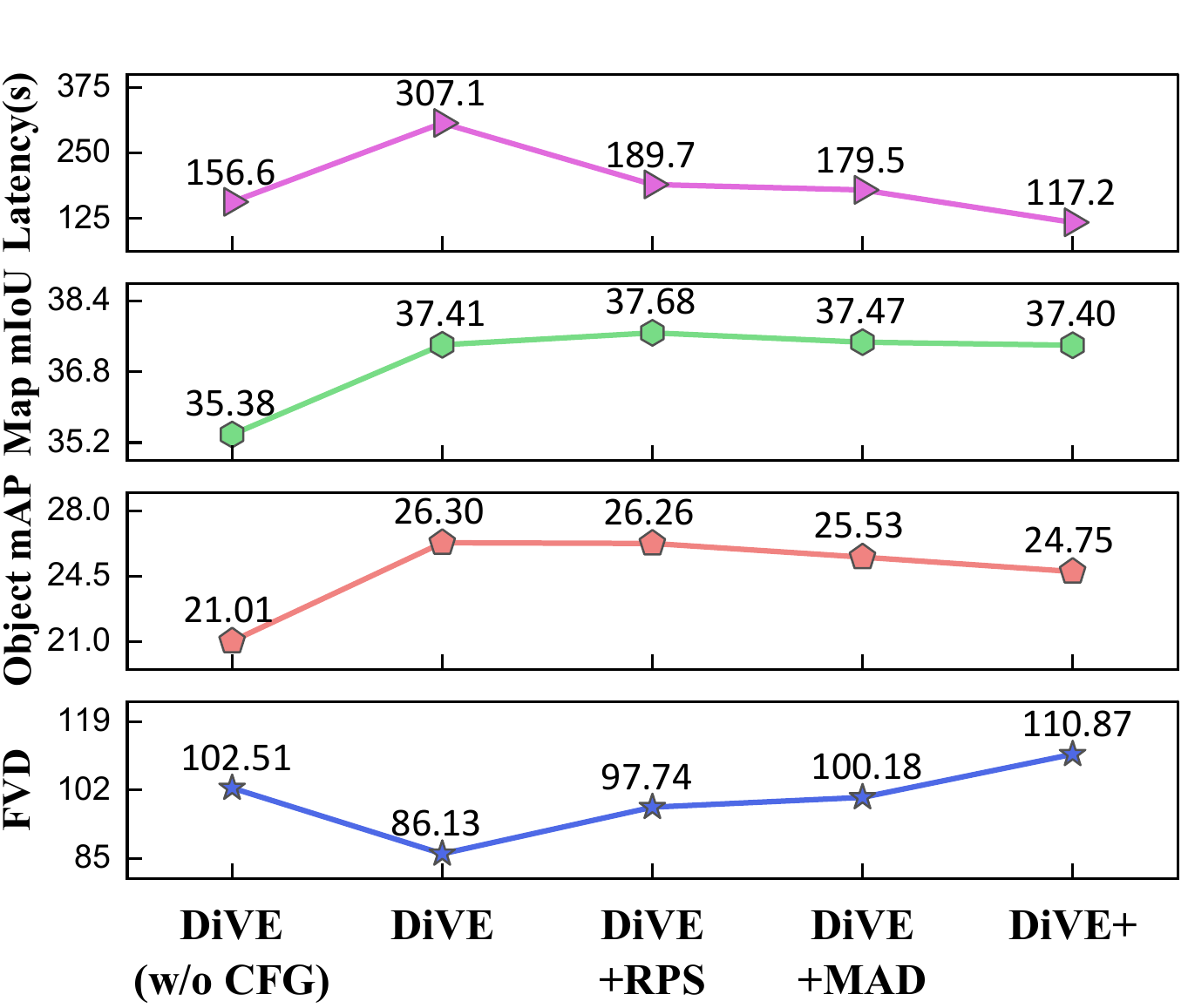}
        \vspace{-0.5cm}
        \caption{DiVE in different settings.}
        \label{fig:accab}
    \end{minipage}%
    \hfill
    \begin{minipage}[t]{0.655\textwidth}
        \centering
        \includegraphics[width=\textwidth]{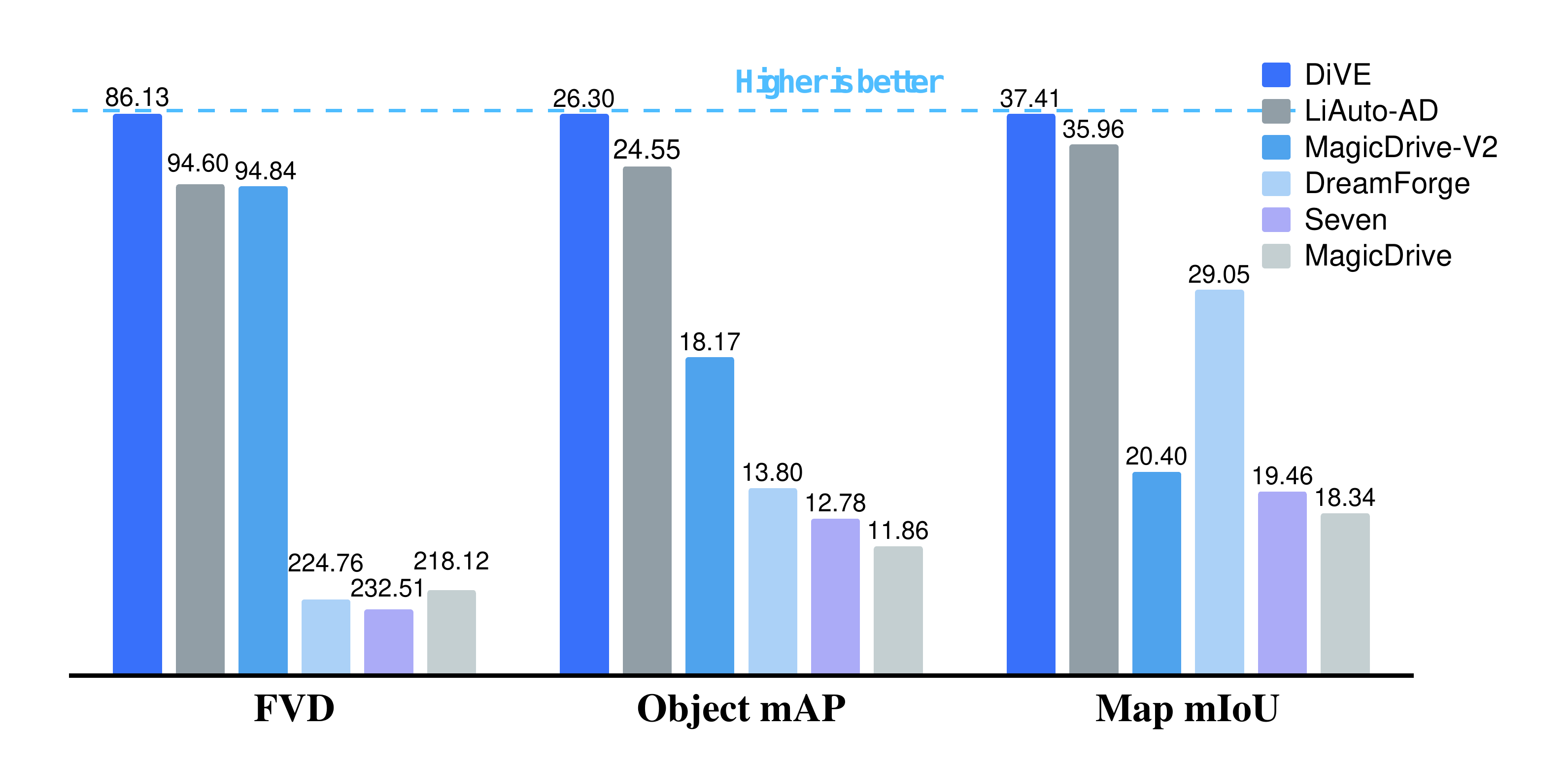}
        \vspace{-0.5cm}
        \caption{DiVE vs. W-CODA baseline~\cite{gaomagicdrive}, winners~\cite{du2024challenge,mei2024dreamforge,jiang2024dive} and MagicDrive-V2~\cite{gao2024magicdrivev2}.}
        \label{fig:workshop_cmp}
    \end{minipage}
    \vspace{-0.3cm}
\end{figure*}

\noindent \textbf{Qualitative Results.}
In Figure~\ref{fig:qualitative_cmp}, DiVE shows superior generation quality compared to MagicDrive~\cite{gaomagicdrive} and Panacea~\cite{wen2024panacea}, which suffer from issues like vehicle fragmentation and unrealistic artifacts (\emph{e.g}., rearview mirrors on vehicle backs). DiVE excels in producing highly realistic vehicles with precise fidelity and maintains both temporal and cross-view consistency, in contrast to the significant color and shape variations in MagicDrive and Panacea. Additionally, DiVE ensures precise scene controllability, generating correct object counts, road layouts, and realistic zebra crossings, while supporting flexible customization of weather, time, architectural styles, and even vehicle colors (Figure~\ref{fig:edit} and Figure~\ref{fig:control})---a capability often absent in prior methods. Notably, whether MAD and RPS are applied individually or jointly to enhance inference efficiency, the visual quality of DiVE's outputs remains nearly indistinguishable from non-accelerated generation, thereby validating the effectiveness of these acceleration strategies.

\noindent \textbf{Generated Videos for Data Augmentation.}
To evaluate whether DiVE-generated data can enhance perception tasks, we adopt an evaluation protocol analogous to Panacea by synthesizing a new training dataset with DiVE and assessing its impact on StreamPETR~\cite{wang2023exploring}. Experimental results demonstrate that When trained on the DiVE-generated dataset, the model achieves mAP and NDS scores of 27.0 and 40.4, respectively---corresponding 71.1\% and 82.4\% of the performance attained using purely real-world data (Table~\ref{tab:2}). This confirms the viability of synthetic data as a valuable training resource. Moreover, augmenting the original dataset with DiVE-generated data further boosts StreamPETR's performance across all metrics, with mAP and NDS gains exhibiting substantially larger margins compared to those achieved by Panacea and Glad~\cite{xie2025glad}. These findings underscore DiVE's superiority in augmenting perception systems through high-fidelity synthetic data generation.

\subsection{Ablation Study}
\label{subsec:abl}
Given the high resolution and frame count of DiVE-generated videos, full validation inference demands substantial resources. To streamline evaluation of Sec.~\ref{subsec:abl} and Sec.~\ref{subsec:more_comp}, we adopt W-CODA's approach, generating only the first 16 frames per scene across four runs, assessing quality via FVD and controllability using BEVFormer~\cite{li2022bevformer} for 3D object detection and BEV segmentation.

\noindent \textbf{View-Inflated Attention.} We conduct an ablation study comparing view-inflated attention with two baseline mechanisms: (1) self-attention (which lacks view interaction) and left-self-right attention (focusing on neighboring views). As shown in Table~\ref{tab:3}, view-inflated attention excels in both generation quality and controllability. While left-self-right attention improves local consistency, it fails to maintain global semantic coherence, thereby underscoring the advantages of view-inflated attention for ensuring cross-view consistency and preserving long-term generation quality.

\noindent \textbf{Multi-Control Auxiliary Branch Distillation (MAD).} To evaluate the efficacy of MAD's Auxiliary Branch and Mixed-Control Guidance Training, we perform ablation studies (Table~\ref{tab:mad}) demonstrating their superiority over alternative methodologies. For the Auxiliary Branch, comparisons with DICE's Enhancer~\cite{zhou2025dice} and PnP's ControlNet~\cite{hsiao2024plug} reveal critical limitations: Enhancer exhibits suboptimal multi-condition performance due to restricted parameter capacity, while ControlNet---despite marginally higher effectiveness---suffers from computational inefficiency and an inflexible single-guidance-scale design, compromising multi-dimensional control balance. Furthermore, Mixed-Control Guidance Training outperforms single-condition distillation strategies across all metrics (implementation details in Algorithm~\ref{alg:ab1} and~\ref{alg:ab2}), establishing its advantage in stabilizing multi-condition guidance learning.

\noindent \textbf{Resolution Progressively Sampling (RPS).} Table~\ref{tab:4} validates the effectiveness of the resolution-aware timestep shift in RPS and demonstrates the rationality of timestep distribution across resolutions. The $i$-$j$-$k$ notation denotes the number of sampling steps at 240p, 360p, and 480p resolutions, respectively. Without timestep shift (\emph{e.g}., 10-10-10), performance degrades consistently, underscoring its importance. While increasing steps at higher resolutions improves quality, it also escalates inference time; thus, 10-10-10 strikes the optimal balance between performance and efficiency.

\subsection{More Results}
\label{subsec:more_comp}
\noindent \textbf{DiVE's Results in Different Settings.} Figure~\ref{fig:accab} provides a quantitative analysis of DiVE's performance across configurations with and without CFG and under MAD/RPS. Independent application of MAD and RPS achieves 1.71$\times$ and 1.62$\times$ inference acceleration, respectively, with slight degradation in FVD and Object mAP. Notably, Map mIoU exhibits marginal improvement on smaller sample sizes, corroborating the efficacy of both techniques. Crucially, their synergistic implementation yields shorter inference latency than CFG-disabled DiVE while achieving stronger controllability despite marginally reduced FVD scores, thereby demonstrating their mutual compatibility for high-efficiency video generation.

\noindent \textbf{Quantitative Results in W-CODA'24.} Figure~\ref{fig:workshop_cmp} compares DiVE’s performance against existing methods under the W-CODA evaluation framework. DiVE consistently outperforms both UNet-based~\cite{gaomagicdrive,du2024challenge,mei2024dreamforge} and DiT-based~\cite{jiang2024dive,gao2024magicdrivev2} models across all metrics, with particularly notable advancements in generative quality and fine-grained controllability.

%% file: section/5_conclusion.tex
\section{Conclusion}

We present DiVE, a pioneering DiT-based framework for generating multi-view driving scene videos. Through various improvements to DiT architecture, DiVE is capable of generating videos that precisely align with 3D annotations and maintain temporal and multi-view consistency. To further accelerate the inference in DiVE, we propose two strategies: Multi-Control Auxiliary Branch Distillation and Resolution Progressive Sampling. The former addresses the challenge of CFG selection while achieving superior inference efficiency without compromising performance. The latter eliminates the need for persistent high-resolution inference, maintaining generation quality while enhancing efficiency in a training-free manner. Both DiVE and its variant DiVE+ (integrated with the MAD and RPS) achieve SOTA performance. With generating higher-quality videos efficiently, DiVE stands out as the preferred choice for enhancing data used in training perception tasks of autonomous driving. 

%% file: section/appendix.tex
\section{More Related Work}
\subsection{Diffusion Transformer for Generation}
Recently, Diffusion Transformer (DiT)~\cite{peebles2023scalable} has emerged as a remarkable breakthrough by replacing the U-Net backbone with a Vision Transformer architecture~\cite{dosovitskiy2021an}, enhancing the scalability and efficiency of Latent Diffusion Models (LDMs)~\cite{rombach2022high} in various generative tasks. SD3~\cite{esser2024scaling} is one of the pioneering works that applied DiT to the text-to-image generation, utilizing rectified flow~\cite{liuflow} and introducing new noise samplers to achieve better performance. Owing to the success of DiT in image generation, recent research has focused on exploring its potential in video generation. Sora~\cite{sora} was the first to demonstrate DiT's capability in generating high-fidelity and strongly consistent videos, inspiring open-source frameworks like OpenSora~\cite{opensora} and CogVideoX~\cite{yang2025cogvideox}. OpenSora~\cite{opensora} employs a 3D VAE to construct hybrid representations and models temporal dependencies via 1D temporal attention. However, limited by the representational capacity of 1D attention, CogVideoX~\citep{yang2025cogvideox} adopts 3D attention to jointly capture spatial-temporal relationships. Existing methodologies, however, are constrained to uni-modal text conditioning. In contrast, our task scenario demands simultaneous controllability across four distinct modalities (text , road sketches , 3D object instances , and camera parameters), posing significant challenges in model architecture design and training.

\subsection{High-Resolution Generation}
High-resolution diffusion models seek to overcome fixed resolution limitations. DemoFusion~\citep{du2024demofusion} employs progressive upscaling with noise inversion but requires full inference steps per scale. CheapScaling~\citep{guo2024make} introduces tuning-free pivot replacement mechanism and U-Net-specific time-aware upsampling for multi-scale synthesis. Megafusion~\citep{wu2024megafusion} adopts a tuning-free truncate-and-relay strategy but lacks rectified flow optimization. In contrast to these image-centric approaches, our method leverages DiVE's multi-resolution generation with resolution-aware timestep shift~\citep{esser2024scaling}, enabling efficient high-quality video synthesis through progressive low-to-high resolution sampling while reducing computational overhead.

\subsection{Multi-View and Panorama Generation}
Recent advancements in multi-view video synthesis predominantly focus on 4D asset generation~\cite{xie2025svd,li2024vividzoo} through the synergistic fusion of 3D geometric priors and temporal motion modeling. SV4D~\cite{xie2025svd} generates multi-view video by simultaneously leveraging 3D priors from multi-view image generation models and motion priors from video generation frameworks, guided by a reference video input. Extending this paradigm, Vivid-ZOO~\cite{li2024vividzoo} bridges the modality gap between 3D-consistent image generation and video synthesis through Low-Rank Adaptation (LoRA) layers, effectively resolving domain misalignment. These advances have driven progress in 360° panorama synthesis. 360DVD~\cite{wang2024360dvd} introduces a 360-adapter module to adapt pre-trained video generators to equirectangular projections of panoramic data via fine-tuning. 4K4DGEN~\cite{li2025kdgen} proposes a training-free paradigm that achieves 4K-resolution spherical panorama synthesis through parallel denoising of multiple viewpoints. CubeDiff~\cite{kalischek2025cubediff} decomposes cubemap faces into individual perspective images and employs multi-view generative models for coherent synthesis across all six faces. Despite their advancements, these approaches lack explicit control over BEV (Bird's Eye View) layouts for simulating diverse real-world driving scenarios, limiting their applicability in generating augmented training data for downstream perception systems.

\section{More Implementation Details}
\subsection{Distillation Training Algorithms}

Algorithm~\ref{alg:alg} details the workflow of Mixed-Control Guidance Training, while Algorithm~\ref{alg:ab1} and~\ref{alg:ab2} outline two Single-Control Guidance Training strategies for ablation studies, each differing in implementation mechanisms. The underlined components emphasize their key operational distinctions from Mixed-Control paradigm.
\begin{algorithm}
    \caption{Mixed-Control Guidance Training}
    \label{alg:alg}
    \begin{algorithmic}[1]
        \REQUIRE Dataset $\mathcal{D}$, Pre-trained multi-condition diffusion model $\upsilon_{\theta}(\cdot,\cdot,\cdot,\cdot)$
        \renewcommand{\algorithmicrequire}{\textbf{Initial:}}
        \REQUIRE Model with Auxiliary Branches $\mathbf{\upsilon}_{[\theta,\psi_l,\psi_i,\psi_r]}(\cdot,\cdot,\cdot,\cdot)$
        \WHILE{not converged}
        \STATE Sample video-text-instance-sketch pairs $x_1,\mathcal{L},\mathcal{I},\mathcal{R} \sim \mathcal{D}$
        \STATE Sample a noise $x_0 \sim \mathcal{N}(0,I)$
        \STATE Sample a timestep $t \sim Uniform[0,1]$
        \STATE Sample a guidance scale $\omega \sim Uniform[1,8]$
        \STATE $x_t=(1-t)x_0+tx_1$
        \STATE Set control conditions: $[l, i, r] \in$ \\ $\left( \{\phi, \mathcal{L}\} \times \{\phi, \mathcal{I}\} \times \{\phi, \mathcal{R}\} \right) \setminus \left\{ (\mathcal{L}, \mathcal{I}, \mathcal{R}) \right\}$
        \STATE $\omega_l=\omega_i=\omega_r=\omega$
        \STATE $\upsilon'_\theta=(\omega+1)\upsilon_\theta(x_t,\mathcal{L},\mathcal{I},\mathcal{R}) - \omega \upsilon_\theta(x_t,l,i,r)$
        \STATE \textbf{Define active parameters:}
        \STATE $\Psi = \{\psi_\diamond \mid \diamond \in \{l,i,r\}, \diamond \neq \phi\}$
        \STATE $\Omega=\{\omega_\diamond \mid \diamond \in \{l,i,r\}, \diamond \neq \phi\}$
        \STATE $L(\Psi) = \lVert \mathbf{\upsilon}_{[\theta,\Psi]}(x_t,\mathcal{L},\mathcal{I},\mathcal{R},\Omega) - \upsilon'_\theta \rVert^2_2$
        \STATE $\Psi \leftarrow \Psi - \eta \nabla_{\Psi}L(\Psi)$
        \ENDWHILE
    \end{algorithmic}
\end{algorithm}

\begin{algorithm}
    \caption{Single-Control Guidance Training Strategy 1}
    \label{alg:ab1}
    \begin{algorithmic}[1]
        \REQUIRE Dataset $\mathcal{D}$, Pre-trained multi-condition diffusion model $\upsilon_{\theta}(\cdot,\cdot,\cdot,\cdot)$
        \renewcommand{\algorithmicrequire}{\textbf{Initial:}}
        \REQUIRE Model with Auxiliary Branches $\mathbf{\upsilon}_{[\theta,\psi_l,\psi_i,\psi_r]}(\cdot,\cdot,\cdot,\cdot)$
        \WHILE{not converged}
        \STATE Sample video-text-instance-sketch pairs $x_1,\mathcal{L},\mathcal{I},\mathcal{R} \sim \mathcal{D}$
        \STATE Sample a noise $x_0 \sim \mathcal{N}(0,I)$
        \STATE Sample a timestep $t \sim Uniform[0,1]$
        \STATE Sample a guidance scale $\omega \sim Uniform[1,8]$
        \STATE $x_t=(1-t)x_0+tx_1$
        \STATE Set control conditions: $[l, i, r] \in$ \\ \underline{$ \left\{ (\phi, \mathcal{I}, \mathcal{R}), (\mathcal{L},\phi,\mathcal{R}), (\mathcal{L},\mathcal{I},\phi) \right\}$}
        \STATE $\upsilon'_\theta=(\omega+1)\upsilon_\theta(x_t,\mathcal{L},\mathcal{I},\mathcal{R}) - \omega \upsilon_\theta(x_t,l,i,r)$
        \STATE \textbf{Define active parameters:}
        \STATE $\Psi = \{\psi_\diamond \mid \diamond \in \{l,i,r\}, \diamond \neq \phi\}$
        \STATE $L(\Psi) = \lVert \mathbf{\upsilon}_{[\theta,\Psi]}(x_t,\mathcal{L},\mathcal{I},\mathcal{R},\omega) - \upsilon'_\theta \rVert^2_2$
        \STATE $\Psi \leftarrow \Psi - \eta \nabla_{\Psi}L(\Psi)$
        \ENDWHILE
    \end{algorithmic}
\end{algorithm}

\begin{algorithm}
    \caption{Single-Control Guidance Training Strategy 2}
    \label{alg:ab2}
    \begin{algorithmic}[1]
        \REQUIRE Dataset $\mathcal{D}$, Pre-trained multi-condition diffusion model $\upsilon_{\theta}(\cdot,\cdot,\cdot,\cdot)$
        \renewcommand{\algorithmicrequire}{\textbf{Initial:}}
        \REQUIRE Model with Auxiliary Branches $\mathbf{\upsilon}_{[\theta,\psi_l,\psi_i,\psi_r]}(\cdot,\cdot,\cdot,\cdot)$
        \WHILE{not converged}
        \STATE Sample video-text-instance-sketch pairs $x_1,\mathcal{L},\mathcal{I},\mathcal{R} \sim \mathcal{D}$
        \STATE Sample a noise $x_0 \sim \mathcal{N}(0,I)$
        \STATE Sample a timestep $t \sim Uniform[0,1]$
        \STATE Sample a guidance scale $\omega \sim Uniform[1,8]$
        \STATE $x_t=(1-t)x_0+tx_1$
        \STATE Set control conditions: $[l, i, r] \in$ \\ \underline{$ \left\{ (\phi, \mathcal{I}, \mathcal{R}), (\phi,\phi,\mathcal{R}), (\phi,\phi,\phi) \right\}$}
        \IF{$l=\phi$}
        \STATE \underline{$\upsilon'_\theta=(\omega+1)\upsilon_\theta(x_t,\mathcal{L},\mathcal{I},\mathcal{R}) - \omega \upsilon_\theta(x_t,l,i,r)$}
        \STATE $L(\psi_l) = \lVert \mathbf{\upsilon}_{[\theta,\psi_l]}(x_t,\mathcal{L},\mathcal{I},\mathcal{R},\omega) - \upsilon'_\theta \rVert^2_2$
        \STATE $\psi_l \leftarrow \psi_l - \eta \nabla_{\psi_l}L(\psi_l)$
        \ELSIF{$l=i=\phi$}
        \STATE \underline{$\upsilon'_\theta=(\omega+1)\upsilon_\theta(x_t,\phi,\mathcal{I},\mathcal{R}) - \omega \upsilon_\theta(x_t,l,i,r)$}
        \STATE $L(\psi_i) = \lVert \mathbf{\upsilon}_{[\theta,\psi_i]}(x_t,\mathcal{L},\mathcal{I},\mathcal{R},\omega) - \upsilon'_\theta \rVert^2_2$
        \STATE $\psi_i \leftarrow \psi_i - \eta \nabla_{\psi_i}L(\psi_i)$
        \ELSE
        \STATE \underline{$\upsilon'_\theta=(\omega+1)\upsilon_\theta(x_t,\phi,\phi,\mathcal{R}) - \omega \upsilon_\theta(x_t,l,i,r)$}
        \STATE $L(\psi_r) = \lVert \mathbf{\upsilon}_{[\theta,\psi_r]}(x_t,\mathcal{L},\mathcal{I},\mathcal{R},\omega) - \upsilon'_\theta \rVert^2_2$
        \STATE $\psi_r \leftarrow \psi_r - \eta \nabla_{\psi_r}L(\phi_r)$
        \ENDIF
        \ENDWHILE
    \end{algorithmic}
\end{algorithm}

\subsection{First-$k$ Frame Masking.}
To enable arbitrary-length video generation, we introduce a first-$k$ frame masking strategy, allowing the model to seamlessly predict future frames from the preceding ones. Formally, given a binary mask $m \in \mathbb{R}^{T}$, where $m_i=1$ for $i\leq k$ and 0 otherwise and the masked frames serve as the condition for future frame generation, we update $x_t$ as
\begin{equation}
x_{t} \leftarrow x_t \odot (1 - m) + x_1 \odot m ~,
\end{equation}
with losses calculated only on unmasked frames. At inference, generation proceeds autoregressively, using the last-$k$ frames as context for seamless extension.

\subsection{Classifier-free Guidance}
We observe that simply nullifying text conditions in the unconditional sampling of Classifier-free Guidance (CFG)~\cite{ho2021classifier} proves ineffective. While this approach often enhances generation fidelity, it leads to blurriness in 3D objects, weakening the model's control over them. Inspired by~\cite{gaomagicdrive}, we extend the unconditional sampling of CFG. Specifically, during unconditional sampling, we simultaneously nullify the text condition $\mathcal{L}$, 3D objects $\mathcal{I}$ and road sketches $\mathcal{R}$ to (\emph{i.e}., set to $\phi$). This jointly optimizes generation fidelity and controllability. The modified velocity estimate is as follows:
\begin{align}
\upsilon'_\theta &= \upsilon_\theta(x_t, \phi, \phi, \phi) \\
&+\lambda\cdot(\upsilon_\theta(x_t, \mathcal{L}, \mathcal{I}, \mathcal{R})-\upsilon_\theta(x_t, \phi,\phi,\phi)) ~,
\end{align}
where $\lambda$ denotes the guidance scale.

\begin{figure}[t]
\centering
\includegraphics[width=\linewidth]{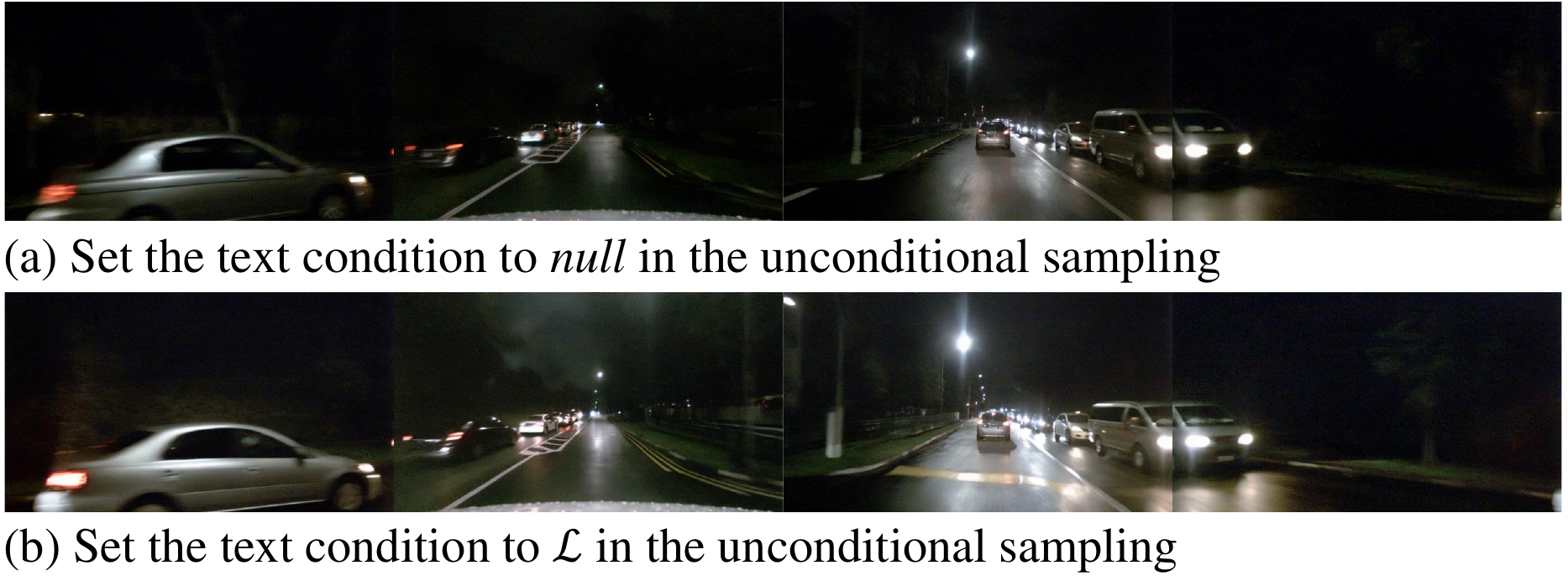}
\caption{Night scene generation in different CFG methods.}
\label{fig:night}
\end{figure}

Notably, night scene generation with ours CFG results in excessive darkness (Figure~\ref{fig:night}a), as DiVE inherently produces reasonably lit nighttime outputs without CFG. However, CFG further emphasizes the darkness, leading to the disappearance of many foreground objects. To address this issue, we retain the text condition (avoiding $\phi$) while nullifying object and sketch inputs during unconditional sampling. This preserves nighttime realism, as shown in Figure~\ref{fig:night}b. The adjusted velocity estimate becomes:
\begin{align}
\upsilon'_\theta &= \upsilon_\theta(x_t, \mathcal{L}, \phi, \phi) \\
&+\lambda\cdot(\upsilon_\theta(x_t, \mathcal{L}, \mathcal{I}, \mathcal{R})-\upsilon_\theta(x_t, \mathcal{L},\phi,\phi)) ~.
\end{align}

\subsection{More Training Details}
All three stages of DiVE are trained with the Brain Floating Point (BF16) precision, and the AdamW optimizer with a learning rate of 1e-4 is adopted. To handle multi-resolution training , we utilize the Bucket mechanism~\cite{chenpixart}. For the primary resolutions of 240p, 360p, and 480p, batch sizes are 4, 2, and 1, respectively. The resolution for StreamPETR~\cite{wang2023exploring} training is 480p, which is different from the 256$\times$704 of the baseline.

\section{Ablation of Camera Information}
\begin{figure}[t]
    \centering
    \includegraphics[width=\linewidth]{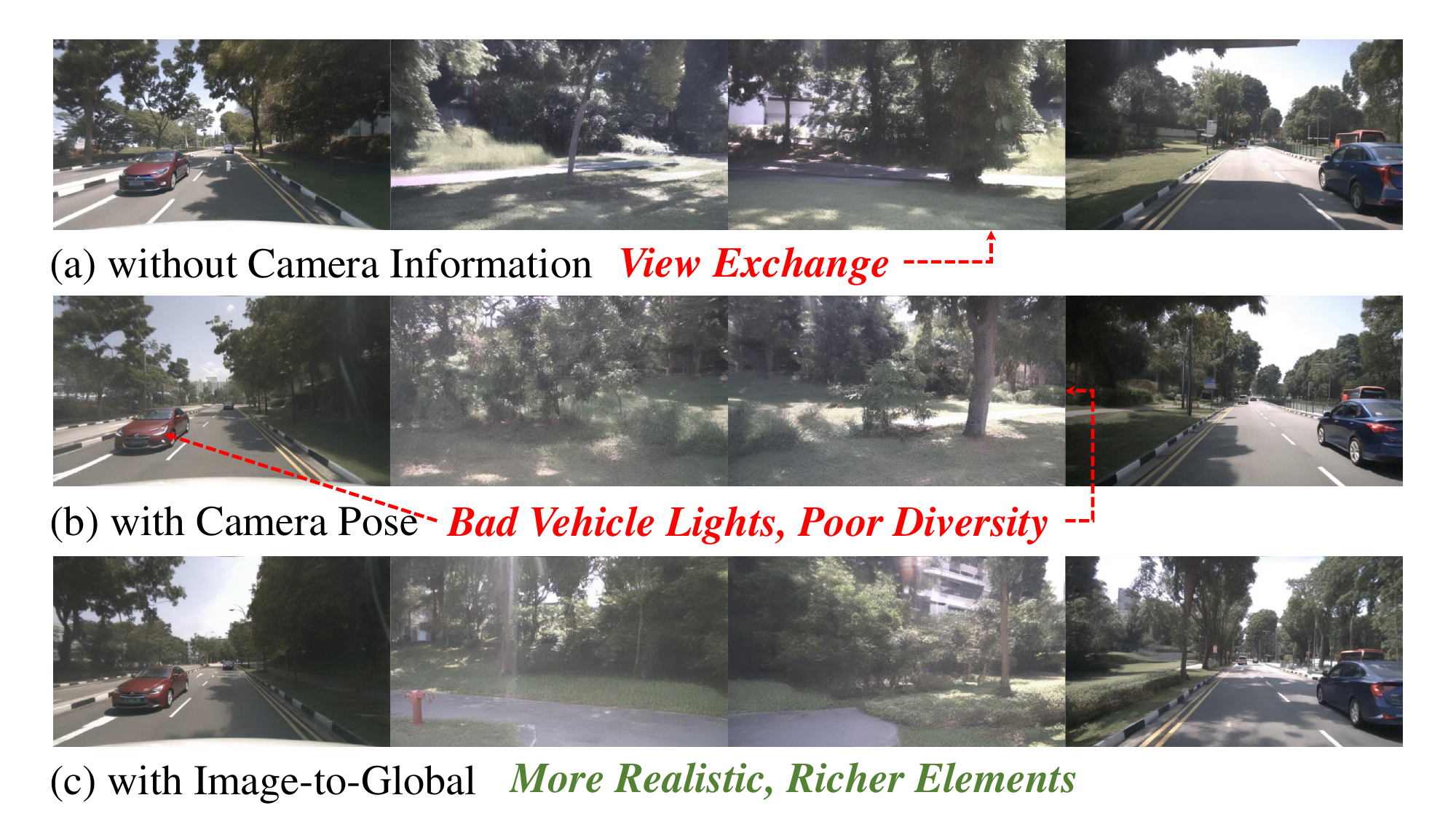}
    \caption{Effect of camera information.}
    \label{fig:camera}
\end{figure}

We find that DiVE can generate multi-view videos with reasonable motion trajectories even without camera information, but this capability lacks reliability. When adjacent views lack both the road sketch and 3D object guidance, the generated video is more likely to exhibit viewpoint exchange issues, as shown in Figure~\ref{fig:camera}a. To address this, we integrate camera information into the model. In contrast to prior methods~\cite{gaomagicdrive} that rely on camera poses, we use an image-to-global coordinate transformation matrix. This design choice is motivated by empirical observations: the image-to-global transformation matrix yields more realistic and element-rich scenes. For example, as shown in Figure~\ref{fig:camera}c, the middle two views include additional elements (\emph{e.g}., fire hydrants and buildings) absent in outputs using camera poses (Figure~\ref{fig:camera}b).

\section{More Visualization Results}
\subsection{Flexible Controllable Generation.}
\begin{figure}[h]
    \centering
    \includegraphics[width=\linewidth]{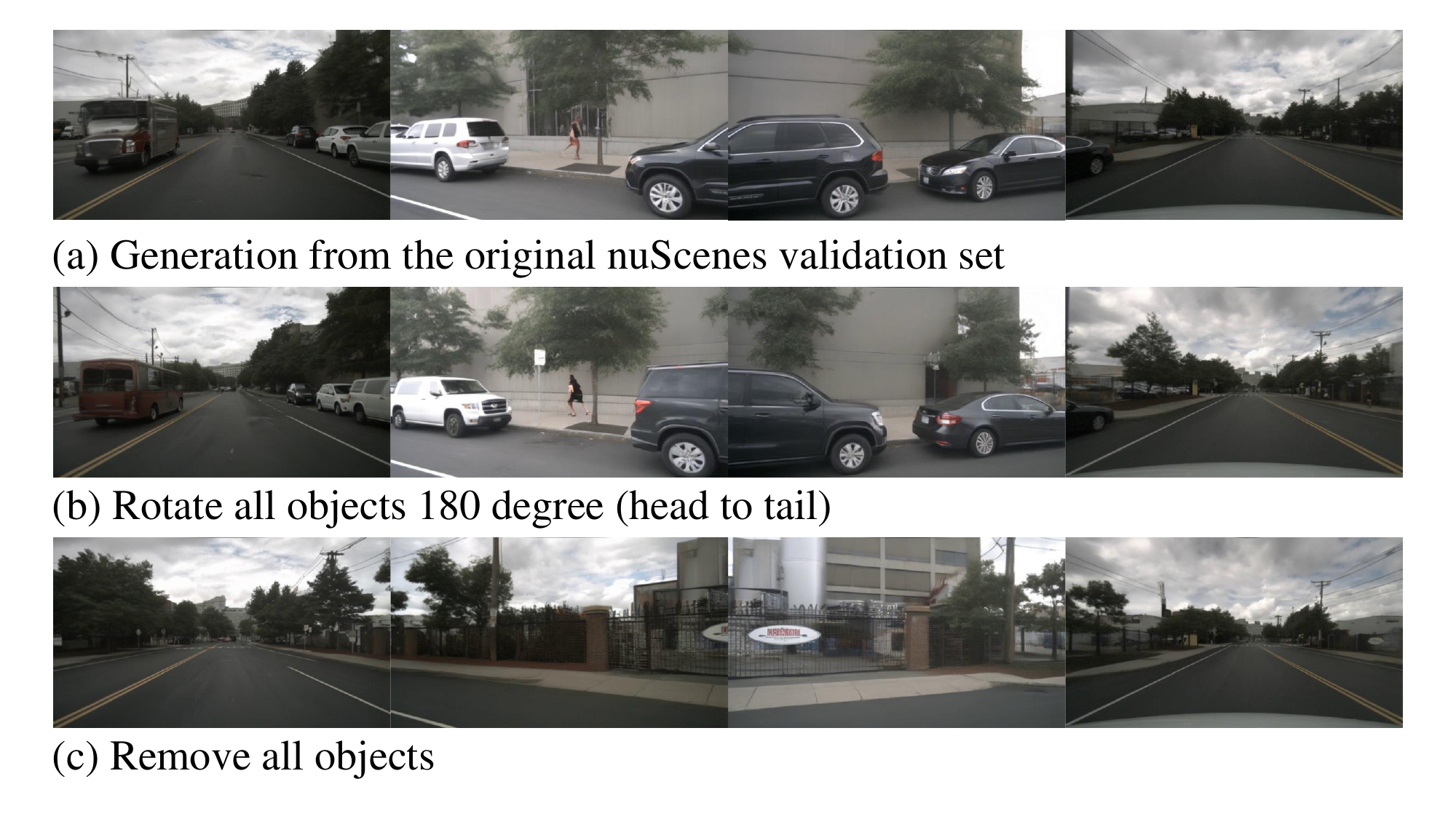}
    \caption{Flexible controllable generation from DiVE.}
    \label{fig:control}
\end{figure}

Figure~\ref{fig:control} demonstrates DiVE’s generation capabilities under varying object conditions. When rotating all the objects by 180°, DiVE retains the capability to produce high-fidelity outputs with precise orientation alignment. In addition to rotation, we also present the generation result after removing all the objects. It is clearly observed that the architecture of the scene has changed, which demonstrates the diverse generation ability of DiVE.

\subsection{More Qualitative Comparison.}
\begin{figure*}[t]
    \centering
    \includegraphics[width=0.90\linewidth]{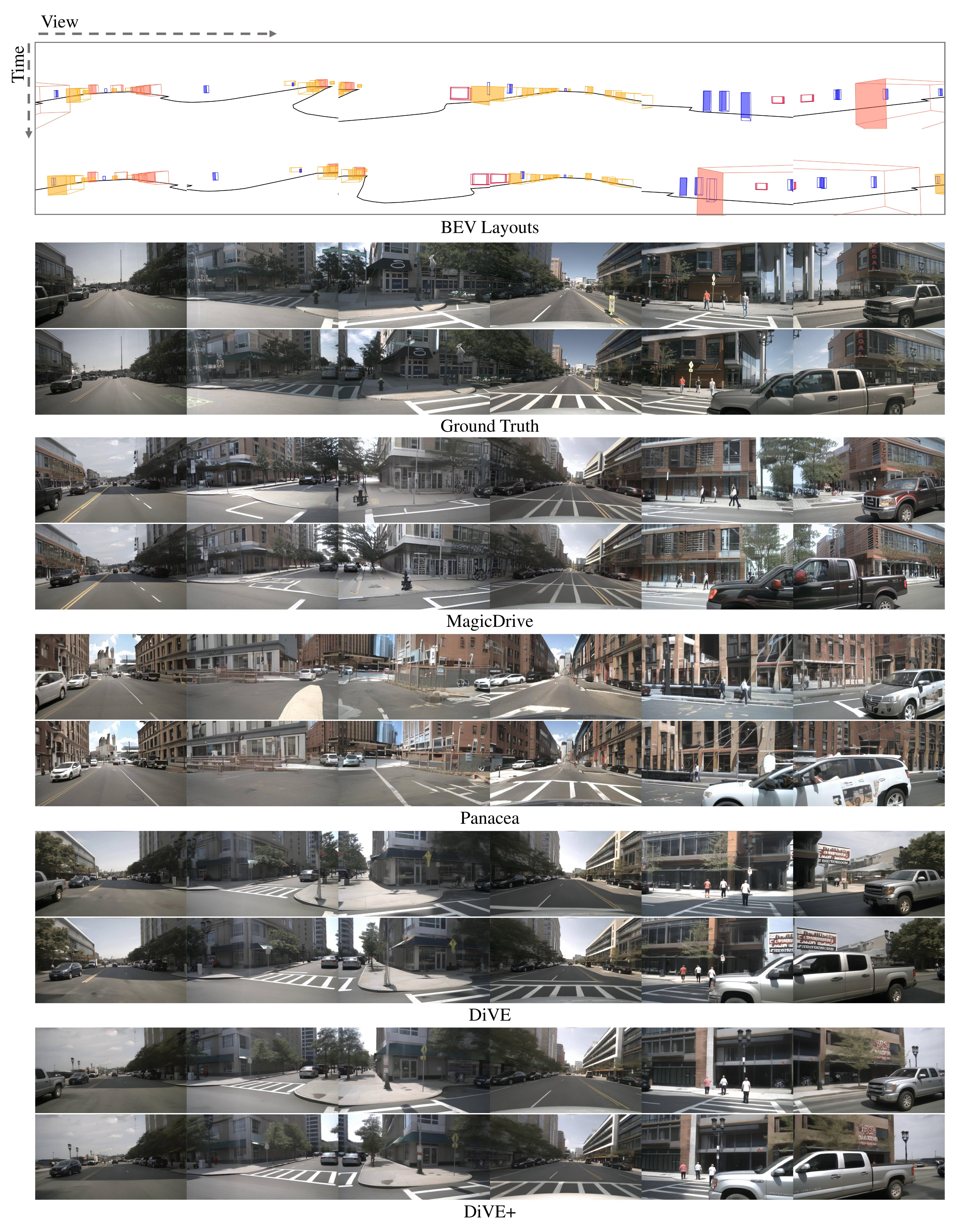}
    \caption{Qualitative comparison with MagicDrive and Panacea on driving scene from nuScenes validation set.}
    \label{fig:comp1}
\end{figure*}

\begin{figure*}[t]
    \centering
    \includegraphics[width=0.90\linewidth]{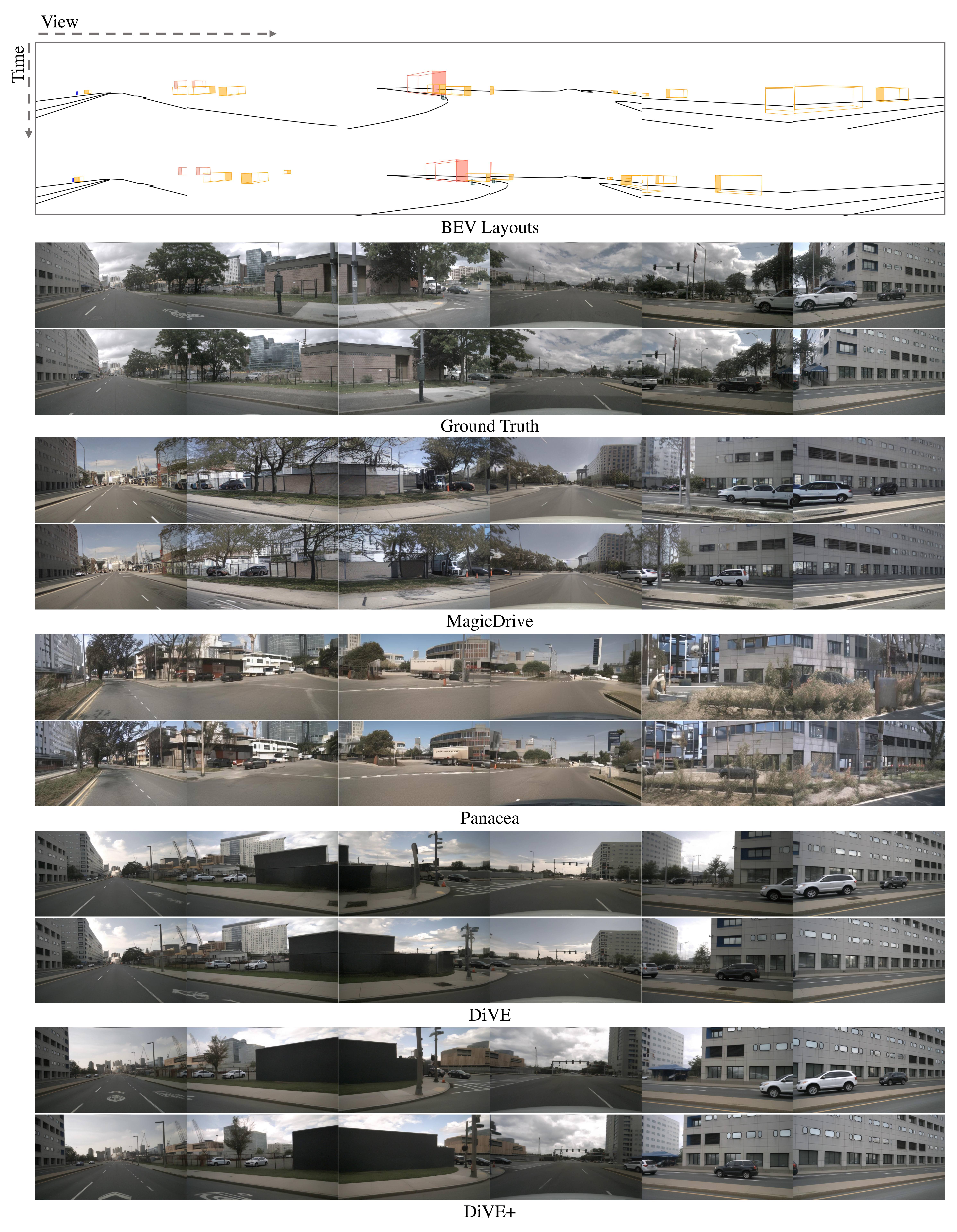}
    \caption{Qualitative comparison with MagicDrive and Panacea on driving scene from nuScenes validation set.}
    \label{fig:comp2}
\end{figure*}

We present more qualitative comparisons with MagicDrive~\cite{gaomagicdrive} and Panacea~\cite{wen2024panacea} in Figure~\ref{fig:comp1},~\ref{fig:comp2}. DiVE consistently demonstrates superior visual fidelity, along with enhanced cross-view and temporal consistency, achieving results that are both natural and realistic.

\subsection{Long Video Generation.}
\begin{figure*}[t]
    \centering
    \includegraphics[width=0.80\linewidth]{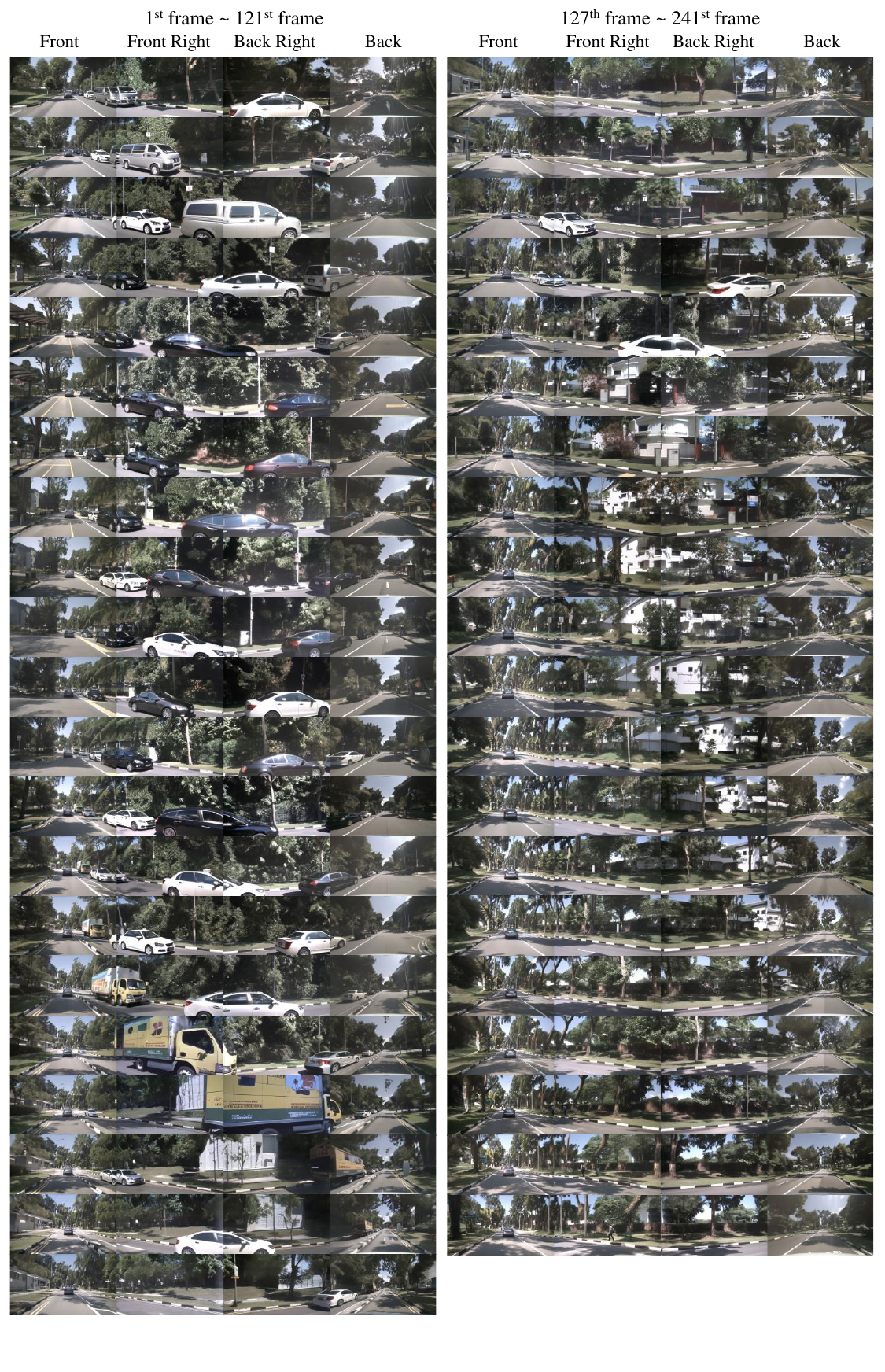}
    \caption{Long video generated by DiVE.}
    \label{fig:long}
\end{figure*}

Figure~\ref{fig:long} illustrates the results of long video generation using DiVE, where each row displays the keyframe outputs. Surprisingly, DiVE exceptional temporal consistency even during the continuous generation of 240 frames, effectively avoiding repetitive patterns or visual artifacts. Moreover, the vehicle color remains unchanged regardless of the temporal sequence or the presence of other vehicles in the scene. Such a high level of consistency is uncommon in prior methods , highlighting DiVE’s advancement in this domain.

\section{Limitations}
Although DiVE can generate driving scenes that are strikingly realistic, the strategies for effectively utilizing generated data as augmentation samples has the potential to yield greater performance gains than solely focusing on advancing generative models. Future work could explore leveraging techniques such as dataset distillation to fully harness the benefits of generated data.

